\documentclass[journal]{IEEEtran} 

\IEEEoverridecommandlockouts                              

\usepackage[utf8]{inputenc}
\usepackage[T1]{fontenc}
\usepackage{nicefrac}
\usepackage{graphicx} 
\usepackage{amsmath,bm} 
\usepackage{amssymb}  
\usepackage{mathrsfs}
\usepackage{times} 
\usepackage{mathtools}
\usepackage{cancel}
\usepackage{supertabular}
\usepackage{epsfig} 
\usepackage{comment} 
\usepackage{algorithm,algorithmic}
\usepackage{empheq}
\usepackage{hyperref}
\usepackage{xcolor}
\newcommand{\rev}[1]{{\color{black} #1}}
\usepackage{marginnote}

\title{3D Underactuated Bipedal Walking via H-LIP based Gait Synthesis and Stepping Stabilization}
\author{Xiaobin Xiong, \IEEEmembership{Member, IEEE} and Aaron Ames, \IEEEmembership{Fellow, IEEE}
       \thanks{Some preliminary results of the paper were presented in IROS 2019 \cite{xiong2019orbit}. 
       The work is supported by Amazon Fellowship in Artificial Intelligence. The authors are with the Department of Mechanical and Civil Engineering, California Institute of Technology. Corresponding author: Xiaobin Xiong ({\texttt{xxiong@caltech.edu}).} The videos of the results can be seen in \rev{\href{https://youtu.be/3FolwqfHFMI}{\textbf{\texttt{youtu.be/3FolwqfHFMI}}}} as well as in \cite{video:sim, video:Directional, video:versatile, video:robust}.
}
}  
\begin{document}
\maketitle

\begin{abstract}
In this paper, we \rev{holistically} present a Hybrid-Linear Inverted Pendulum (H-LIP) based approach for synthesizing and stabilizing 3D \rev{foot-underactuated} bipedal walking, \rev{with an emphasis on thorough hardware realization}. The H-LIP is proposed to capture the essential components of the underactuated and actuated part of the robotic walking. The robot walking gait is then \textit{directly} synthesized based on the H-LIP. We comprehensively characterize the periodic orbits of the H-LIP and provably derive the stepping stabilization via its step-to-step (S2S) dynamics, which is then utilized to approximate the S2S dynamics of the horizontal state of the center of mass (COM) of the robotic walking. The approximation facilities a H-LIP based stepping controller to provide desired step sizes to stabilize the robotic walking. By realizing the desired step sizes, the robot achieves dynamic and stable walking. The approach is fully evaluated in both simulation and experiment on the 3D underactuated bipedal robot Cassie, which demonstrates dynamic walking behaviors with both high versatility and robustness. 
\end{abstract}


\section{INTRODUCTION}


\rev{Bipedal walking robots locomote in the world by actuating its internal joints \cite{raibert1986legged, grizzle2014models}. The foot-ground contact can be underactuated when the foot contacts the ground partially or the foot itself is not internally actuated due to the lack of motors at the ankle joints in the robot design \cite{raibert1986legged, rezazadeh2018robot} for agility and simplicity. Both are forms of foot-underactuation where the ground can not fully and continuously react moments to the robot. 
Moreover, the constant switching of support legs renders the dynamics to be hybrid \cite{grizzle2014models}: cycling between continuous dynamics and discrete transitions. These conditions differ from those of controlling a robot arm \cite{lynch2018modern} where the base is fixed and the dynamics is typically continuous and fully actuated. Thus, generally speaking, it is challenging to control locomotion behaviors on the high dimensional foot-underactuated bipedal walking robots. }

   \begin{figure}[t]
      \centering
      \includegraphics[width = 1\columnwidth]{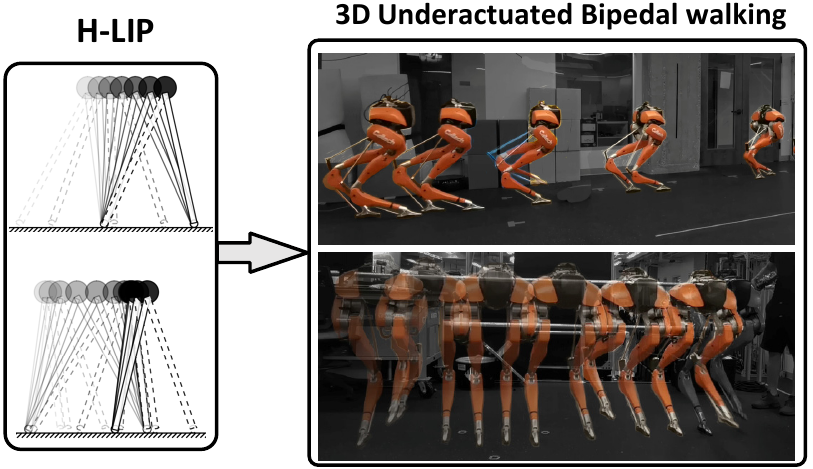}
      \caption{The H-LIP based approach on generating underactuated bipedal walking: (left) the periodic orbits of the H-LIP and (right) 3D robotic walking.} 
      \label{fig:overview}
\end{figure}

 Various approaches have been proposed to generate stable foot-underactuated bipedal walking. The hybrid zero dynamics (HZD) approach \cite{westervelt2003hybrid,grizzle2014models}, born in the control community, plans attractive periodic orbits with virtual constraints in the full-dimensional state-space of the robot via large-scale trajectory/parameter optimizations \cite{hereid20163d} or numerical methods \cite{rosaCont}. Feedback controllers \cite{ames2014rapidly,westervelt2003hybrid} are employed to enforce the virtual constraints, and the stability of the generated walking is typically determined by the analysis on the numerically derived Poincaré map. Practical realizations, however, challenge the theoretical soundness. The optimization is highly non-convex and difficult to solve in general. Furthermore, the stability of optimized 3D walking gaits cannot be easily determined in the optimization. Lastly, even if the optimized gait is stable, the walking on the robot can be unstable if the robot deviates from the model that is used in the optimization. 

 Another widely-studied approach uses the Spring Loaded Inverted Pendulum (SLIP) \cite{blickhan1989spring, raibert1986legged} model for generating compliant legged locomotion behaviors. The SLIP sparked wide interests in the legged locomotion community since it was found to successfully capture both walking and running dynamics of biological systems \cite{geyer2006compliant}. The controllers on the SLIP can be derived either by intuition \cite{raibert1986legged} or based on its return map \cite{wu20133}. Several bipedal robots \cite{ahmadi2006controlled, grimes2012design, rezazadeh2015spring, xiong2018bipedal} have been designed and built to resemble the SLIP. For those types of robots, the SLIP-inspired controllers are thus utilized to render the corresponding locomotion behaviors. However, the SLIP based approaches cannot be easily used for non-SLIP like underactuated bipedal robots in practice. Furthermore, transitions between periodic walking behaviors on both the SLIP and the robot are not well studied.

 In this paper, we present a walking synthesis and control based on a model simplification, which focuses on the switch of support legs and neglects foot actuation. The simplified model is a variant of the Linear Inverted Pendulum (LIP) \cite{kajita2002} with passive pivot contact and a hybrid domain structure. Thus, we name it the Hybrid-LIP (H-LIP) \cite{xiong2019orbit}. The H-LIP is passive in the continuous domains of walking, and the "actuation" that changes the walking behavior is on the step size. By formulating the dynamics at the step-level, the step size becomes the input to its step-to-step (S2S) dynamics.
 
 The H-LIP approximates the hybrid walking of a \rev{foot-underactuated bipedal robot} assuming that the center of mass (COM) is approximately constant and the swing foot periodically lifts off and strikes the ground. Then, the linear S2S dynamics of the H-LIP approximates the S2S dynamics of the robot. By treating the model difference as a bounded disturbance to the linear S2S, state-feedback stepping controllers (i.e., \textit{H-LIP stepping} \cite{xiong2019orbit, xiong2020ral}) can be synthesized to control the horizontal COM state of the robot at the pre-impact event; the difference of the horizontal states between the robot and the H-LIP converges to disturbance invariant sets. 
 
To implement the H-LIP based approach on the 3D robot, we first realize desired walking behaviors on the H-LIP, by characterizing its periodic orbits and synthesizing their stabilization. The desired H-LIP walking is then used in the stepping controller to find desired step sizes on the robot to realize desired walking behaviors. We realize our approach on the 3D underactuated bipedal robot Cassie, shown in Fig. \ref{fig:overview}. The desired walking trajectories are constructed based on the H-LIP and its stepping controller and then stabilized via joint-level controllers. Versatile and robust walking behaviors are thus realized on the robot in both simulation and experiments.

\subsection{Contributions}
The main contributions of this paper are:
\begin{itemize}
    \item \textbf{\rev{Holistically presenting} the low-dimensional model (H-LIP) with its comprehensive orbit characterizations and stabilization \rev{for the purpose of approximating underactuated bipedal walking dynamics.}} We \rev{examine the nature of underactuated bipedal walking and then extend the underactuated canonical LIP to a hybrid version to \textit{approximate} the underactuated bipedal walking}. 
     \item \textbf{\rev{Designing a highly versatile gait synthesis with stepping stabilization directly based on the H-LIP for realizing 3D underactuated walking on robots \textit{with and without compliance}.}} We present a walking synthesis that directly maps the features of the H-LIP walking to the robotic walking. \rev{All desired trajectories are designed in closed-form, eliminating the need of solving any non-convex trajectory optimization problems}.
     The stepping stabilization is formally synthesized based on the S2S dynamics approximation of the robot via the H-LIP.   
    \item \textbf{\rev{Providing} a computationally-efficient and robust realization on the physical hardware of the complex 3D underactuated bipedal robot Cassie with passive compliance.} We present computationally-efficient and rigorous implementations to solve the practical problems including contact detection and COM velocity approximations, which are shown to be highly robust to uncertainties of the hardware system and external disturbances.  
\end{itemize}

\rev{This paper extends on our previous results in \cite{xiong2019orbit, xiong2020ral}. In \cite{xiong2019orbit}, we presented the H-LIP, its orbit characterization, and heuristically-synthesized but provable stepping stabilization. We applied the stepping controller on an actuated SLIP (aSLIP) model and the underactuated bipedal robot Cassie in simulation. Trajectory optimization on the aSLIP is needed to apply the H-LIP based stepping on the aSLIP itself and Cassie. \cite{xiong2020ral} later extends the aSLIP in 3D and embeds its dynamics on fully-actuated humanoid walking; \cite{xiong2021ral} extends the aSLIP walking on rough terrains via advanced control techniques. 

Compared to our previous results, this paper formally presents the H-LIP based gait synthesis and stabilization on foot-underactuated bipedal robots in a holistic and direct fashion, with a very strong emphasis on the comprehensive hardware realization and evaluation on a 3D robot via principled model-based techniques. Importantly, the gait synthesis is newly designed to directly map the actuated states from the H-LIP to the robot; constructed outputs are capable to address the compliance in the robot, which can also be applied to the robot without compliance. Versatile walking behaviors can be easily realized in the new gait synthesis directly without solving any trajectory optimization problems on the robot or the aSLIP. The direct gait synthesis also improves the S2S approximation and thus the robustness to external disturbances. Besides these, the equivalent characterizations of the Period-1 and Period-2 orbits are proved to complete the orbit identification. The stepping controllers are rigorously derived, and different low-level QP controllers are shown to work equivalently for output tracking. The performance of the realized walking on the 3D robot is analyzed on the error S2S dynamics. }


\subsection{Related Work \rev{and Comparisons}}
The H-LIP is a variant of the canonical LIP model \cite{kajita2001} with foot-underactuation and hybrid dynamics. The LIP has been extensively applied in the Zero Moment Point (ZMP) approach \cite{kajita2003biped, feng2016robust} for realizing humanoid walking. The LIP is continuously actuated; the H-LIP is only discretely actuated by swapping support legs. One can view the ZMP-LIP approaches as using the ZMP of the LIP to approximate the ZMP of the robot with the LIP dynamics directly embedded on the humanoid. Instead, we use the H-LIP dynamics to approximate the horizontal COM dynamics of foot-underactuated bipedal robots, which can not strictly embed the pendulum dynamics. Additionally, compared to the LIP with foot-placement controllers \cite{pratt2012capturability, feng2016robust, griffin2017walking} on humanoid walking, this approach focuses on the periodic walking and S2S stabilization on foot-underactuated bipedal robots.

Compared to the periodic walking realized via HZD \cite{grizzle2014models, gong2018feedback, li2020animated}, the periodic orbits of the H-LIP are directly controlled via the step size on the S2S dynamics. Thus, the stability of the orbits and their transitions are solely determined by the stepping controller. Additionally, the robot is not necessarily controlled to evolve on a strict orbit in its state-space. Instead, it converges closely to the walking behavior of the H-LIP. The H-LIP walking is pre-determined but the walking of the robot is not. 

The notation of the S2S dynamics in legged locomotion is an adaptation of the Poincaré return map in nonlinear dynamics \cite{khalil1996noninear}. The S2S has mostly appeared in controlling SLIP running \cite{bhounsule2020approximation, GeyerChapter}. By investigating the evolution of the apex states, the S2S/return map of running can be easily obtained on the SLIP. Feedback controllers thus can be synthesized based on the S2S to stabilize the running of the SLIP. However, the S2S of the walking of the SLIP has not been shown to be obtained easily, possibly due to the complexity of the inclusion of the DSP dynamics. Similarly, the S2S of a 3D bipedal walking robot cannot be obtained easily. By and large, the control based on the return map of walking has been focused on the linearization at the fixpoint \cite{bhounsule2014low, kuo1999stabilization,bhounsule2015discrete, wensing2013high} of a periodic solution on the return map (very few exceptions \cite{bhounsule2020approximation, morimoto2007improving} learned the S2S); \rev{heuristically tuned foot-placement controllers \cite{raibert1986legged, rezazadeh2015spring} implicitly exploit the S2S dynamics for stabilization around local behaviors. This paper, instead, approximates the robot S2S dynamics over a large region in the state-space at the Poincaré section. Additionally, the S2S approximation is linear and readily facilitates periodic walking to be completely characterized and feedback controllers to be rigorously designed, which yields versatile and robust gait synthesis.}

\section{Preliminary} 
\label{sec:prelim}
\subsection{Hybrid Dynamics Model of Bipedal Walking}
The dynamics of bipedal walking can be described as a hybrid dynamical system \cite{grizzle2014models} with continuous dynamics in different domains and in-between discrete transitions. The continuous dynamics are affine control systems: 
\begin{equation}
    \dot{x} = f_v(x) + g_v(x) \tau,
\end{equation}
where $x$ is the system state, $\tau$ is the vector of input torques, and the subscript $_v$ is the domain index. The discrete transitions between the consecutive domains can be described by:
\begin{equation}
    x^+ = \Delta_{v \rightarrow v+1}(x^-),
\end{equation}
where the superscripts $^-$ and $^+$ stand for the instants before and after the transition, respectively.

\begin{figure}[b]
    \centering
    \includegraphics[width = .9\columnwidth]{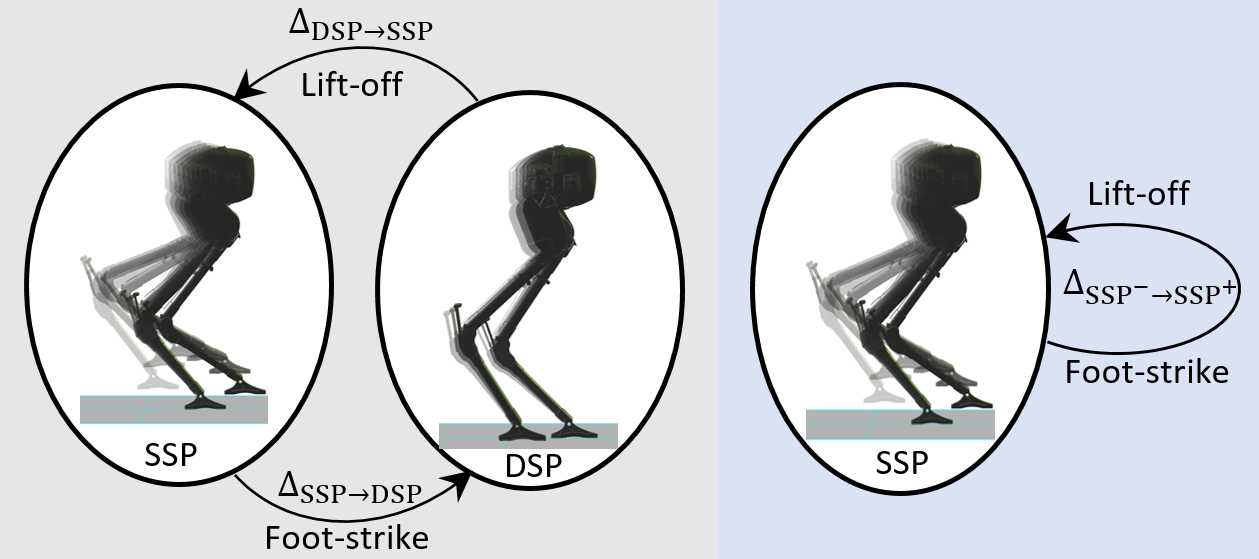}
    \caption{The hybrid graphs of one-domain walking and two-domain walking.}
    \label{fig:hybridgraph}
\end{figure}
\rev{Since there is no flight phase,} the hybrid walking dynamics are composed either of a single domain: single support phase (SSP) or by two domains: a SSP and a double support phase (DSP). We refer to the two as one-domain walking and two-domain walking. 
 For one-domain walking, the transition $\Delta_{ \text{SSP}^-\rightarrow \text{SSP}^+}$ happens at the impact when the swing foot strikes the ground. The impact is modeled as plastic impact \cite{grizzle2014models} where the velocity of the swing foot becomes zero after the impact, and thus the state undergoes a discrete jump. As for the two-domain walking, the transition $\Delta_{ \text{SSP} \rightarrow \text{DSP}}$ is also the impact event, and the transition $\Delta_{\text{DSP} \rightarrow \text{SSP}}$ is when one of the stance feet lifts off from the ground. The existence of the DSP happens when there is compliance in the leg that prevents the stance leg from instantaneously lifting off at the impact. The one-domain walking can be viewed as a two-domain walking with an instantaneous DSP. The two-domain walking is then chosen as the general model that we study for walking.
 
\begin{figure}[t]
    \centering
    \includegraphics[width = 0.6\columnwidth]{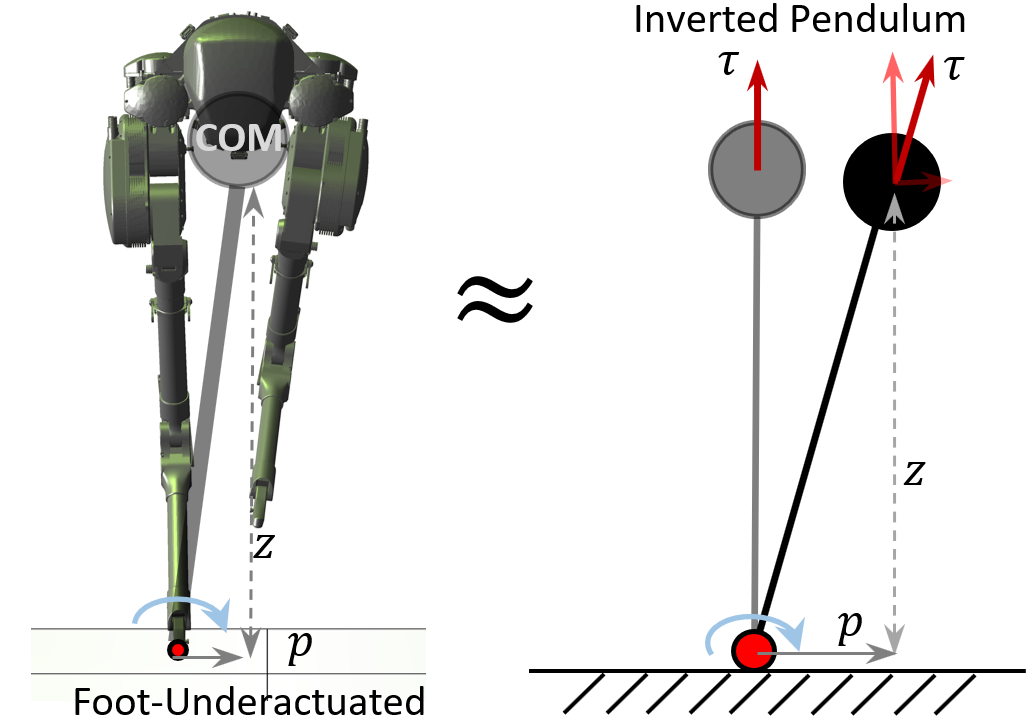} 
    \caption{\rev{The actuated and underactuated states of bipedal walking with the illustration by an inverted pendulum in the SSP.}} 
    \label{fig:underactuation}
\end{figure}

\subsection{\rev{Actuated and Underactuated States}}

The foot-underactuation prevents the direct continuous control on the horizontal center of mass (COM) state of the robot to desired trajectories in the SSP. A simple illustration is that an inverted pendulum would roll passively without any actuation at the contact with the ground (see Fig. \ref{fig:underactuation}). \rev{Like the robot, if we assume there is an actuation on the pendulum length, the vertical COM height of the pendulum $z$ can be chosen as the actuated state, while the horizontal COM position $p$ becomes the underactuated state.} Theoretically speaking, the robot has rotational linkages, which can change its centroidal angular momentum and thus indirectly affect the horizontal COM \cite{pratt2012capturability, kajita2003resolved, xiong2020sequential}, e.g., the angle of a flywheel-inverted pendulum can be controlled via the continuous rotation of the flywheel \cite{Flywheel}. However, the joints on robots typically have limited ranges of motion, control bandwidth, and torques in practice; \rev{the robot also needs to move the swing leg in a certain fashion to maintain walking}. Thus, it is not possible to purely depend on the angular momentum to continuously control the COM. Therefore, the horizontal COM state of the underactuated robot is approximately equivalent to the underactuated ("weakly actuated" \cite{da2019combining}) states in practice, \rev{which we can not continuously control to follow random continuous trajectories. The rest degrees of freedom, e.g., the vertical COM height and the swing foot position, are the actuated states that we can directly control to follow reasonable trajectories. The way of stabilizing the underactuated state is revealed in the next section by approximating the hybrid underactuated dynamics via the H-LIP.}

\section{Hybrid Linear Inverted Pendulum Model}
\label{sec:HLIP}

\subsection{Walking Dynamics of the H-LIP}
\subsubsection{Hybrid Dynamics}
\begin{figure}[t]
      \centering
      \includegraphics[width = 1\columnwidth]{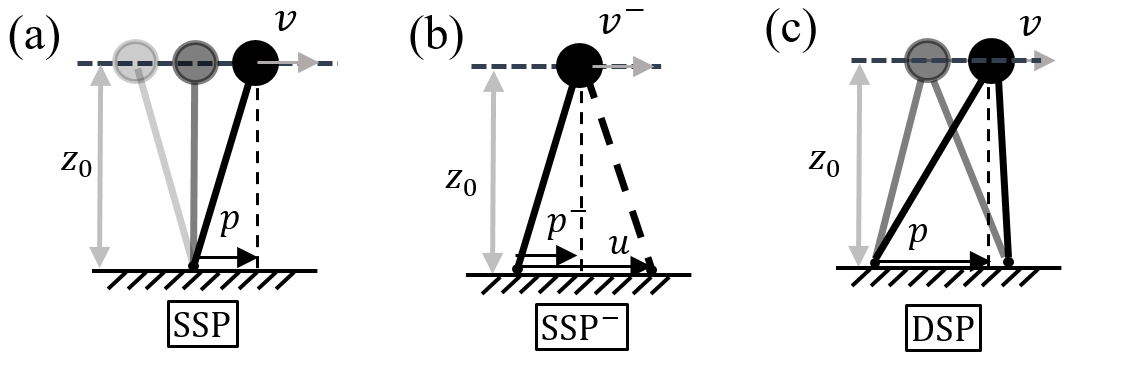}%
      \caption{The walking of the Hybrid-LIP model in SSP (a), at the preimpact state (b), and during DSP (c).}
      \label{fig:H_LIP}
\end{figure}

The H-LIP is a point-mass model with a constant center of mass (COM) height and two telescopic legs with point-feet (see Fig. \ref{fig:H_LIP}). The point-feet correspond to the underactuated feet of bipedal robots. Based on the number of contacts with the ground, the walking is composed by a Single Support Phase (SSP) and a Double Support Phase (DSP). In the SSP, the model is a passive LIP with no actuation; in the DSP, we assume that the mass velocity is constant. The state of the system is composed of the position $p$ and the velocity $v$ of the mass. $p$ is defined as the position of the mass relative to its stance foot. In the DSP, the stance foot is the previous stance foot in the SSP. Thus, the dynamics are:
\begin{align}
 \ddot{p} &= \lambda^2 p,   \tag{SSP}  \\
 \ddot{p} &= 0,   \tag{DSP}
\end{align}
where $\lambda = \sqrt{\frac{\text{\rev{g}}}{z_0}}$ and $z_0$ is the height of the H-LIP. We assume that the domain durations ($T_\text{SSP}$ and $T_\text{DSP}$) are constant. Since the H-LIP is a point-mass model, the swing foot behavior of lift-off and touch-down is not explicitly described. The transitions between domains are assumed to be smooth:
\begin{eqnarray}
 \Delta_{\textrm{SSP} \rightarrow \textrm{DSP}} : \left \{\begin{matrix}
v^{+} = v^{-}  \\
p^{+} = p^{-}
\end{matrix}\right. \quad
\Delta_{\textrm{DSP} \rightarrow \textrm{SSP}} : \left \{\begin{matrix*}[l]
v^{+} = v^{-}  \\
p^{+} = p^{-}   - u
\end{matrix*}\right. \nonumber
\end{eqnarray}
where $u$ is the step size, and the $+/-$ indicate the states after and before the transition, respectively. Since the dynamics are linear and the transitions are in closed-form, the solutions are:
\begin{eqnarray}
\label{eq:LIPsol}
\textrm{SSP} &:& \left \{\begin{matrix*}[l]
p(t)= c_1 e^{\lambda t} + c_2 e^{- \lambda t}  \\
v(t) = \lambda ( c_1 e^{\lambda t} - c_2 e^{- \lambda t})
\end{matrix*}\right.
\\
\label{eq:LIPDSPsol}
\textrm{DSP} &:& \left \{\begin{matrix*}[l]
p(t) = p^{-}_{\textrm{SSP}} + v^{-}_{\textrm{SSP}} t \\
v(t) = v^{-}_{\textrm{SSP}}
\end{matrix*}\right.
\end{eqnarray}
where
$
    c_1 = \frac{1}{2} ( p^{+}_{\textrm{SSP}}+\frac{1}{\lambda}v^{+}_{\textrm{SSP}})$ and $ 
    c_2 = \frac{1}{2} ( p^{+}_{\textrm{SSP}}-\frac{1}{\lambda}v^{+}_{\textrm{SSP}}).
$

\noindent{\underline{\textit{3D H-LIP:}}} The H-LIP is a planar model. Similar to LIP, the H-LIP can be presented in the 3-dimensional space. Since its dynamics are completely decoupled in each plane, a H-LIP in 3D is equivalent to two orthogonally-coupled planar H-LIPs. 

\noindent{\underline{\textit{Equivalence to a One-Domain System:}}}
The hybrid dynamics of the H-LIP with two domains can be equivalently simplified to a single-domain hybrid system. This will simplify the descriptions of periodic orbits. Since the closed-form solution of the DSP is known, we virtually treat the DSP and its associated transitions as a single transition from the final state of the SSP to the initial state of the next SSP. Thus, the transition is defined as:
\begin{eqnarray}
\label{eq:ImpactS2S}
 \Delta_{\textrm{SSP}^- \rightarrow \textrm{SSP}^+} &:& \left \{\begin{matrix*}[l]
v^{+} &= v^{-}  \\
p^{+} &= p^{-} +v^{-}T_\text{DSP} - u.
\end{matrix*}\right. 
\end{eqnarray}
As a result, we have a hybrid dynamical system with a continuous SSP dynamics and a virtual discrete transition. When $T_\text{DSP} = 0$, the dynamics becomes an actual one-domain system with only SSPs, which is the passive LIP (LIP with point foot) in the literature \cite{razavi2015restricted, gong2020angular, luo2018self, koolen2012capturability}.

\rev{\noindent{\underline{\textit{Actuated and Underactuated States:}}} The assumptions on the H-LIP are designed to approximate the hybrid dynamics on the foot-underactuated bipedal robot. We include the DSP in the model to make it general to represent both one-domain walking and two-domain walking on the robot. The actuated states of the H-LIP are implicitly defined on the vertical COM height and the swing foot positions, which match the actuated states of the robot. The assumption of the constant COM height is to simplify the dynamics, which will be enforced on the robot. The swing foot position changes the step size $u$ at a fixed stepping frequency. The contact is unactuated to match the foot-underactuation. The horizontal COM states $[p, v]^T$ of the H-LIP are passive in both domains of walking; however, they are underactuated by the swing foot trajectories via $u$ in the hybrid dynamics, the closed-form relation of which is revealed in the following step-to-step dynamics.}

\begin{figure*}[t]
      \centering
      \includegraphics[width = 0.9\textwidth]{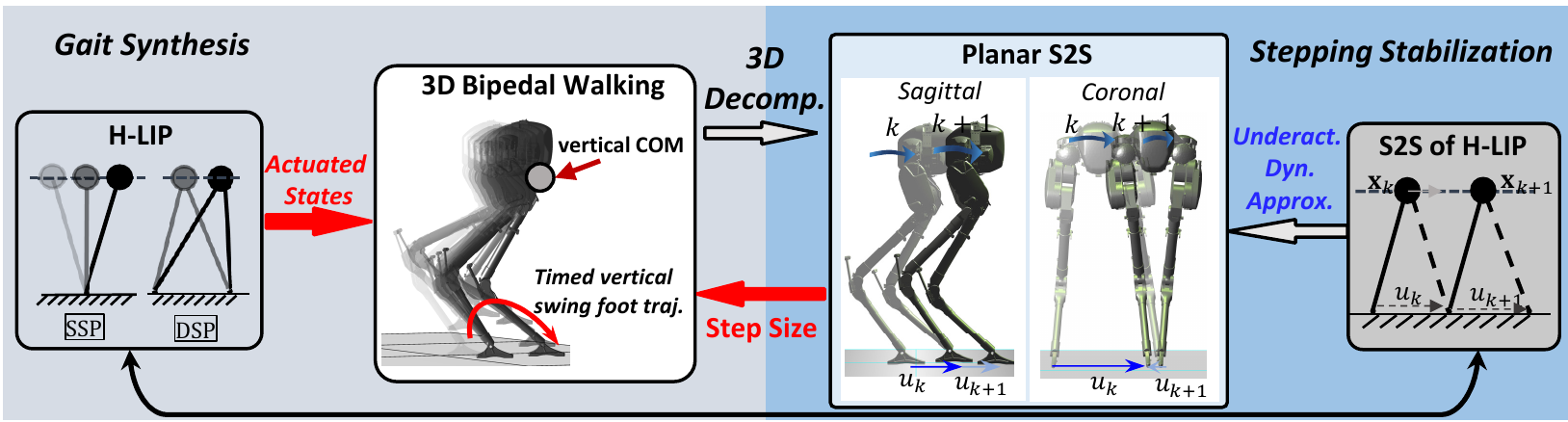}
      \caption{\rev{Illustration of the H-LIP based gait synthesis on the actuated states and the stepping stabilization on the underactuated states of the robot.}}
      \label{fig:gaitSynthesis}
\end{figure*}

\subsubsection{Step-to-step Dynamics} 
The dynamics of the H-LIP are piecewise linear. As the durations are constant, the pre-impact states at consecutive steps can be related in closed-form. The state-space representation of the SSP dynamics is:
\begin{equation}
\label{eq:LIPsspDynamics}
\underset{\dot{\mathbf{x}}_\text{SSP}}{\underbrace{\frac{d}{dt}\begin{bmatrix}
 p \\
 v
\end{bmatrix}}}=  \underset{A_{\textrm{SSP}} }{\underbrace{\begin{bmatrix}
 0 & 1 \\
 \lambda^2 & 0
\end{bmatrix}}} \underset{\mathbf{x}_\text{SSP}}{\underbrace{\begin{bmatrix}
 p \\
 v
\end{bmatrix}}}.
\end{equation}
Thus, the final state of the SSP is calculated from the initial state of the SSP:
\begin{equation}
\label{eq:SSPsol}
\mathbf{x}^{-}_{\text{SSP}_{\textrm{H-LIP}}} = e^{A_{\text{SSP}} T_{\text{SSP}}} \mathbf{x}^{+}_{\text{SSP}_{\textrm{H-LIP}}}.
\end{equation}
The transition in \eqref{eq:ImpactS2S} can be written as:
\begin{equation}
\label{eq:DSPsol}
{\mathbf{x}^{+}_{\text{SSP}}}_{k+1} =
\begin{bmatrix}
1 & T_{\text{DSP}} \\
0 & 1
\end{bmatrix} { \mathbf{x}^{-}_{\text{SSP}}}_{k} + \begin{bmatrix}
-1 \\
0
\end{bmatrix} u_k,
\end{equation}
where $k$ is the step index. Plugging \eqref{eq:DSPsol} into \eqref{eq:SSPsol} yields,
\begin{equation}
\label{eq:LIP-LTI}
{\mathbf{x}^{-}_{\text{SSP}}}_{k+1} = \underset{A} {\underbrace{ e^{A_{\text{SSP}} T_{\text{SSP}}} \begin{bmatrix}
1 & T_{\text{DSP}} \\
0 & 1
\end{bmatrix}}}{ \mathbf{x}^{-}_{\text{SSP}}}_{k} +
\underset{B} {\underbrace{e^{A_{\text{SSP}} T_{\text{SSP}}} \begin{bmatrix}
-1 \\
0
\end{bmatrix}}} u_k. \nonumber
\end{equation}
From now on, we treat the final state of the SSP as the discrete state of the hybrid dynamics of the H-LIP. Thus we drop some subscripts and superscripts and rewrite the above equation as: 
\begin{empheq}[box=\fbox]{align}
\label{eq:HLIP_S2S}
\rev{\mathbf{x}^\text{H-LIP}_{k+1} = A \mathbf{x}^\text{H-LIP}_k + B u^\text{H-LIP}_k}
\end{empheq}
which is referred to as the \textbf{\textit{step-to-step (S2S)} dynamics} of the H-LIP. The S2S is a discrete linear time-invariant system with the step size being the input, \rev{which concisely shows the underactuated state is actuated at the discrete step-level by the actuated state under the hybrid walking dynamics.}


\subsection{H-LIP based \rev{Direct} Gait Synthesis}
\rev{The S2S dynamics of the H-LIP will be used for approximating the underactuated hybrid walking dynamics of the bipedal robot. In order to facilitate the approximation, we need to enforce the \textit{actuated states} of the robot to behave similarly to the actuated states of the H-LIP (Fig. \ref{fig:gaitSynthesis}), including the vertical COM and swing foot trajectories: }


\noindent{\underline{Vertical COM Height}:} The vertical height of the COM $z_\text{COM}$ should remain constant during walking. \rev{Strictly enforcing $z_\text{COM}$ to be constant yields the robot underactuated dynamics \cite{powell2016mechanics, gong2020angular, dai2021bipedal} to be tightly approximately by the H-LIP dynamics in the SSP.} For underactuated robots with passive compliance in the leg (e.g. Cassie), strictly enforcing this condition is challenging; hence, we only make sure that $z_\text{COM}$ is approximately constant on the robot. 

\noindent{\underline{Vertical Swing Foot Trajectory}:} The second component is on the synthesis of the vertical trajectory of the swing foot $z_\text{sw}$. As the step frequency on the H-LIP is assumed to be constant, the swing foot on the robot is expected to periodically lift off and strike the ground with the same frequency. This creates continuing hybrid execution on the dynamical system and makes sure that the S2S dynamics of the robot exists. As a result, $z_\text{sw}$ should evolve on a time-based trajectory, which creates the lift-off and touch-down behaviors based on time.

\noindent{\underline{Horizontal Swing Foot Trajectory}:} 
As the step size is the control input on the H-LIP, the horizontal trajectory of the swing foot should be constructed to achieve a certain desired step size on the robot. Since the impact is time-based, the horizontal trajectory of the swing foot is constructed to swing to the desired step location at the time of impact. 

\noindent{\underline{\textit{3D Walking Decomposition}}:} The application to 3D robotic walking requires an orthogonal composition of two planar H-LIPs. \rev{The two H-LIPs are synchronized with the same domain durations and vertical COM height}. We select the sagittal plane and coronal plane of the robot as the decomposition of the robotic walking. The horizontal COM state, swing foot position, and the step size of the robot are decoupled into those in the sagittal and coronal plane, respectively.

\subsection{Stepping Stabilization via S2S Dynamics Approximation}
\rev{As we use the S2S dynamics of the H-LIP to approximate the S2S dynamics of the robot, we can synthesize the desired step sizes via the S2S dynamics approximation for stabilizing the underactuated horizontal COM state of the robot to realize the desired walking.} The stepping in the sagittal plane is used as an example; it is applied identically to the coronal plane. 

As the robot is controlled to periodically lift off and touch down the foot, the hybrid dynamics of walking repeats a walking cycle. In other words, the pre-impact state exists, despite the number of domains in the hybrid walking. Let $\{q^-, \dot{q}^- \}$ be the pre-impact state of the robot. The step-level evolution of the pre-impact states, i.e., the S2S dynamics of the robot, can be represented by:
\begin{equation}
    \{q^-_{k+1}, \dot{q}^-_{k+1} \} = \mathcal{P}(q^-_k , \dot{q}^-_k,  \tau(t)), \label{eq:robotFullS2S}
\end{equation}
where $k$ is the index of the step, and $\tau(t)$ represents the torques which are applied during the step $k$. Each step starts with a DSP (if exists) and a following SSP. Let $\mathbf{x}^R = [p^R, v^R]^T$ be the horizontal COM state at the pre-impact event. $p^R, v^R$ are the horizontal position and velocity of the COM of the robot, which are functions of the preimpact state $\{q^-, \dot{q}^- \}$. Thus \rev{$\mathbf{x}^R_{k+1} = [p^R_{k+1}(q^-_{k+1}, \dot{q}^-_{k+1}), v^R_{k+1}(q^-_{k+1}, \dot{q}^-_{k+1})]^T$. Plugging \eqref{eq:robotFullS2S} in this yields the S2S dynamics of the horizontal COM state $\mathbf{x}^R_{k+1}$, which can be compactly represented by:}
\begin{equation}
\label{eq:robotS2S}
    \mathbf{x}^R_{k+1} = \mathcal{P}_{\mathbf{x}}(q^-_k , \dot{q}^-_k,  \tau(t)).
\end{equation}
In the latter, we directly refer \eqref{eq:robotS2S} as the S2S of the robot. 

Due to the nonlinear dynamics of the robot, the exact expression of the S2S dynamics can not be computed in closed-form. Thus, synthesizing the controller directly based on the S2S dynamics is difficult in general. Since we design the gait of the robot based on the H-LIP, the S2S of the robot should be close to the S2S of the H-LIP. Therefore, we use \eqref{eq:HLIP_S2S} to approximate \eqref{eq:robotS2S}, which can be rewritten as:
\begin{align}
\label{eq:RobotS2Sapprox}
\mathbf{x}^R_{k+1} &= A \mathbf{x}^R_{{k}}  +  B u^R_{k} + w_k\\
 w_k :&= \mathcal{P}_{\mathbf{x}}(q^-_k , \dot{q}^-_k,  \tau(t))- A \mathbf{x}^R_{{k}}  -  B u^R_{k}.
\end{align}
where $u^R_k$ is the step size of the robot, and $w$ is the difference of the S2S dynamics between the robot and the H-LIP. $w$ is also the integration of the difference of the continuous dynamics over one step between the two systems. As the gait of the robot is designed to match the walking of the H-LIP, the dynamics error should be small. Given a step frequency, each step happens in a finite time (determined by the vertical trajectory of the swing foot). Therefore, $w$, the integration of the continuous error dynamics over a finite time, is assumed to belong to a bounded set, i.e., $w\in W$. $w$ is treated as the disturbance to the linear system. Thus, on the robot, we apply the \textbf{\textit{H-LIP based stepping}}: 
 \begin{equation}
 \label{eq:HLIPstepping}
u^R = u^{\text{H-LIP}}_\rev{k} + K (\mathbf{x}^R_{k} - \mathbf{x}_{k}^{\text{H-LIP}} )
\end{equation}
where $u^{\text{H-LIP}}_\rev{k} $ is the step size on the H-LIP to realize desired walking behaviors, and $K \in \rev{\mathbb{R}^2}$ is the feedback gain to make $A+BK$ stable ($\texttt{eig}(A+BK)<1$). Let $\mathbf{e} = \mathbf{x}^R - \mathbf{x}^\text{H-LIP}$ be the \textit{error state}. Applying \eqref{eq:HLIPstepping} in \eqref{eq:RobotS2Sapprox} yields the \textit{error S2S dynamics}: 
\begin{align}
\label{eq:errorDynamicsHLIPstepping}
    \mathbf{e}_{k+1} = (A+BK) \mathbf{e}_{k}  + w_k.
\end{align}
Since $A+BK$ is stable, the error dynamics has a minimum disturbance invariant set $E$. By definition,
\begin{equation}
    (A+BK) E  \oplus  W \in E.
\end{equation} 
where $\oplus$ is the Minkowski sum. We call $E$ the error invariant set, i.e., if $\mathbf{e}_k \in E$ then $\mathbf{e}_{k+1} \in E$. If $W$ is small, then $E$ is small. \rev{Since the robot kinematics is bounded by its joint limits, its COM position and step sizes are bounded in finite-sized sets ($u^R \in U$ and $p^R \in X_p$) with a pre-specified $z_\text{COM}$. Given a fixed step frequency, the realizable horizontal velocity $v^R$ is also bounded; thus $\mathbf{x}^R\in X$. The desired behavior of the H-LIP then should satisfy $u^\text{H-LIP} \in U \ominus K E$ and $\mathbf{x}^\text{H-LIP} \in X \ominus E$.} Thus, the desired walking behavior (of the horizontal COM state) can be first realized on the H-LIP, and then applying the H-LIP based stepping yields the behavior to be approximately realized on the robot, with the error being bounded by $E$. \rev{In the next section, we present the identification and realization of the desired periodic walking on the H-LIP in terms of $\mathbf{x}_k^\text{H-LIP}$ and $u_k^\text{H-LIP}$ that will be used in \eqref{eq:HLIPstepping} for generating the desired step sizes on the robot for walking realization.}  

\section{Orbit Characterization, Composition and Stabilization on the H-LIP }
\label{sec:HLIPorbit}
We will briefly present the resulting theorems of the orbit characterization and leave the proofs in the Appendix. \rev{The superscripts $^\text{H-LIP}$ are omitted in this section for conciseness.}

\subsection{Orbit Characterization}
The periodic orbits of the H-LIP that encode walking can be geometrically characterized in its state space. We categorize the orbits of interest into two types, Period-1 (P1) and Period-2 (P2) orbits, depending on the number of steps that the orbit contains. P1 orbits have a period of one step, and P2 orbits have a period of two steps. There are also P$N$ ($N>2$) orbits, and we do not investigate them in this paper. 

\begin{figure}[b]
      \centering
      \includegraphics[width = 0.85\columnwidth]{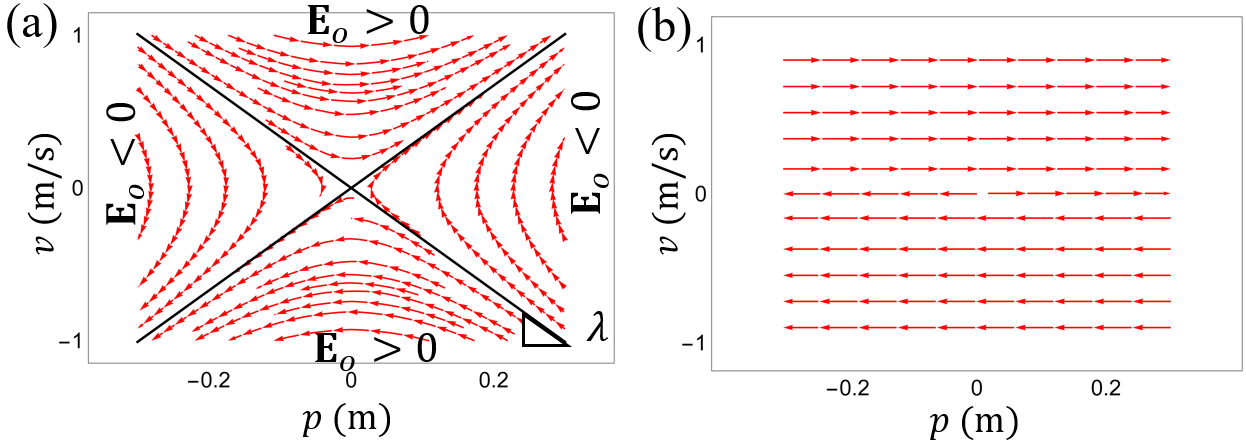}%
      \caption{Phase portraits of the H-LIP walking in its (a) SSP and (b) DSP.}
      \label{fig:phase}
\end{figure}

The H-LIP is a two-dimensional system, thus we can present the periodic orbits explicitly in its state space with its phase portraits. For the H-LIP in SSP, its phase portraits are identical to that of the canonical passive LIP (Fig. \ref{fig:phase} (a)). It is divided into four regions by the cross lines $v = \pm \lambda p$, based on the orbital energy \cite{kajita2001}:
$
\mathbf{E}_o(p, v) = v^2 - \lambda^2 p^2.
$
The physical meaning of $\mathbf{E}_o>0$ is that the H-LIP rotates over the stance foot, i.e., the system passes through the states where $p = 0$. In DSP, the phase portrait is simple, shown in Fig. \ref{fig:phase} (b). 

For conciseness, we use the equivalent one-domain system in Section \ref{sec:HLIP}-A of the H-LIP. Then the orbits can be represented only by a continuous SSP trajectory and a discrete transition. In the following, we present the geometric characterization of P1 and P2 orbits in the phase portrait of the SSP. The subscripts of $_\text{SSP}$ on the states are omitted. Additionally, the pre-impact states and the step sizes of the orbits are presented explicitly from the desired walking velocity.

\begin{figure}[t]
      \centering
      \includegraphics[width = .85\columnwidth]{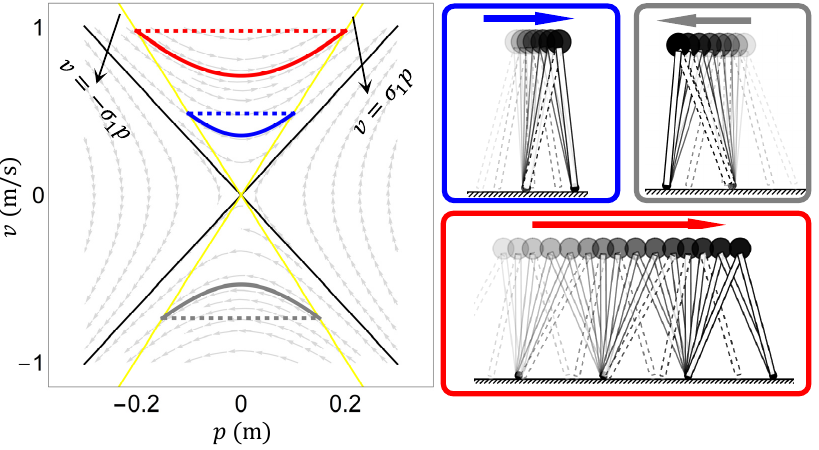}%
      \caption{The Period-1 orbits illustrated in the phase portrait. The yellow cross lines are the orbital lines of P1 orbits. The red, blue and gray lines are walking orbits with $v^- = 1, 0.5, -0.7$ m/s. The dashed lines indicate the transitions in \eqref{eq:ImpactS2S}. The right-side sub-figures illustrate the walking of each orbit.}
      \label{fig:LIP_p1}
\end{figure}

\subsubsection{Period-1 Orbits}
We start with the geometric characterization of the P1 orbits. The velocity is the same between the start and the end of the SSP of the P1 orbits, i.e., $v^+_{\text{SSP}} = v^-_{\text{SSP}}$. Since the phase portrait is left-right symmetric, the orbits are left-right symmetric as well. By inspection, all P1 orbits should only exist in the $\mathbf{E}_o>0$ regions and pass the vertical line of $p = 0$. The comprehensive characterization is stated as follows:
\\

\noindent\fbox{%
    \parbox{\linewidth-6pt}{%
        \textbf{Theorem 1.} The initial and final states of the P1 orbits in SSP are on the \textit{orbital lines} $v = \pm \sigma_1 p$, with
\begin{equation}
\label{eq:sigma1}
\sigma_1: =  \lambda  \text{coth}(\frac{T_\text{SSP}}{2} \lambda),
\end{equation}
being the \textit{orbital slope}. Each state on $v =  \sigma_1 p$ represents the final state of the SSP of a unique P1 orbit with the step size: 
\begin{equation}
\label{eq:u*P1}
u_1 = 2 p^{-} +  T_{\textrm{DSP}}v^{-}.
\end{equation}
    }%
}
\\

Here, we defined the orbital lines and orbital slopes to locate the boundary states of the orbits. Additionally, given a desired net velocity $v^d$, there is a unique P1 orbit for realization. It is obvious that the step size of the P1 orbit is: 
\begin{equation}
\label{P1nominal}
    u^* = v^d(T_\textrm{DSP} + T_\textrm{SSP}): = v^d T.
\end{equation}
with $T$ being the entire step duration. Then the final states of SSP of the P1 orbit are calculated from \eqref{eq:sigma1} and \eqref{eq:u*P1}:
\begin{equation}
\label{eq:desiredP1state}
    [p^*, v^*]  = [1, \sigma_1] \frac{ v^dT }{2 + T_\textrm{DSP} \sigma_1}.
\end{equation}

Fig. \ref{fig:LIP_p1} shows three P1 orbits in the phase portrait of the SSP to illustrate the characterization of the P1 orbits with $T_{\text{SSP}} = 0.5$s. A different $T_{\text{SSP}}$ would produce a different set of orbital lines (the cross yellow lines). As $T_{\text{SSP}} \rightarrow \infty$, the orbital lines converges to the black lines. 

\begin{figure}[t]
      \centering
      \includegraphics[width = .85\columnwidth]{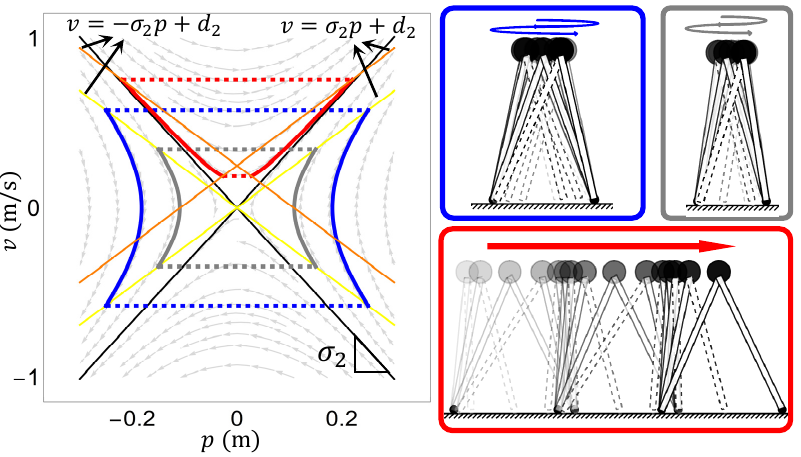}%
      \caption{The Period-2 orbits illustrated in the phase portrait. The yellow cross lines and the \rev{orange} cross lines are the orbital lines of P2 orbits. The blue and gray orbits are the P2 orbits with net velocity being 0. The net velocity of the red orbit is $0.25$m/s. }
      \label{fig:LIP_p2}
\end{figure}

\subsubsection{Period-2 Orbits} 
P2 orbits take two steps to complete a periodic walking. We differentiate the consecutive two steps by its stance foot, indexed by $\mathrm{L/R}$. Similar to the P1 orbits, we identify the orbital slope and orbital lines of P2 orbits, and therefore the P2 orbits are geometrically characterized: 
\\

\noindent\fbox{%
    \parbox{\linewidth-6pt}{%
        \textbf{Theorem 2.} For P2 orbits, the initial and final states of SSP are located on the orbital lines defined as $v = \pm \sigma_2 p + d_2$ with $d_2$ being a constant and the orbital slope:
       \begin{align}
\label{sigma_2}
        \sigma_2 := \lambda \text{tanh}(\frac{T_\text{SSP}}{2} \lambda).
\end{align}
Each state on the line $v = \sigma_2 p + d_2$ represents the final state of the SSP of a P2 orbit with the step size being: 
\begin{equation}
\label{eq:stepLength2}
u_\text{L/R} =2 p^{-}_\text{L/R} +  T_{\textrm{DSP}} v^{-}_\text{L/R}.
\end{equation}
    }%
}
\\

Geometrically, $d_2$ shifts the set of orbital lines up or down. The magnitude of $d_2$ determines the net velocity of the P2 orbit. Given the desired velocity $v^d$, $d_2$ can be calculated:
\begin{align}
\label{eq:d_2}
 d_2 = \frac{\lambda^2 \textrm{sech}^2 (\frac{\lambda}{2}T_{\textrm{SSP}}) T v^{d}}{\lambda^2 T_{\textrm{DSP}} + 2\sigma_{2}},
\end{align}
which does not depend on the selection of the boundary states. This indicates that there is an infinite number of P2 orbits to realize one desired net velocity. Another way to look at this is through the fact that the step sizes are determined by $v^d$: $
    u^*_\text{L} + u^*_\text{R} = 2 v^d T.
$
There are infinite combinations of $u^*_\text{L}, u^*_\text{R}$ to satisfy this, and therefore there are infinite P2 orbits to realize the desired velocity. Selecting one step size (e.g. $u^*_\text{L}$) determines the other one and thus determines the P2 orbit. The final states of the SSP can then be determined from \eqref{eq:stepLength2}, 
\begin{equation}
\label{eq:desiredP2pos}
    p^*_\text{L/R} = \frac{u^*_\text{L/R} - T_{\textrm{DSP}} d_2}{2 + T_{\textrm{DSP}} \sigma_2}, \quad v^*_\text{L/R} = \sigma_2 p^*_\text{L/R} + d_2.
\end{equation}
Fig. \ref{fig:LIP_p2} illustrates three P2 orbits. The blue and the gray orbits are located on the same set of orbital lines (yellow lines with $d_2 = 0$), thus they have a zero net velocity. The red P2 orbit has a non-zero net velocity.

\subsubsection{Equivalent Characterization}
The P1 and P2 orbits are characterized by their orbital lines, respectively. We find that under certain conditions, the orbital lines of P1 orbits can also characterize P2 orbits and vice versa. When $u_\text{L} = u_\text{R}$, a P2 orbit becomes an equivalent P1 orbit, which is stated as: 
\\

\noindent\fbox{%
    \parbox{\linewidth-6pt}{%
        \textbf{Proposition 3.} The orbital lines $v = \pm \sigma_{2} p +d_2$ characterize the P1 orbits when $u_\text{L} = u_\text{R}$, which yields the final state of the SSP as 
$
p^*= \frac{d_2 \textrm{sinh}(T_\textrm{SSP} \lambda)}{2 \lambda}, v^* = \sigma_2 p^* + d_2.
$
    }%
}
\begin{figure}[t]
      \centering
      \includegraphics[width = .75\columnwidth]{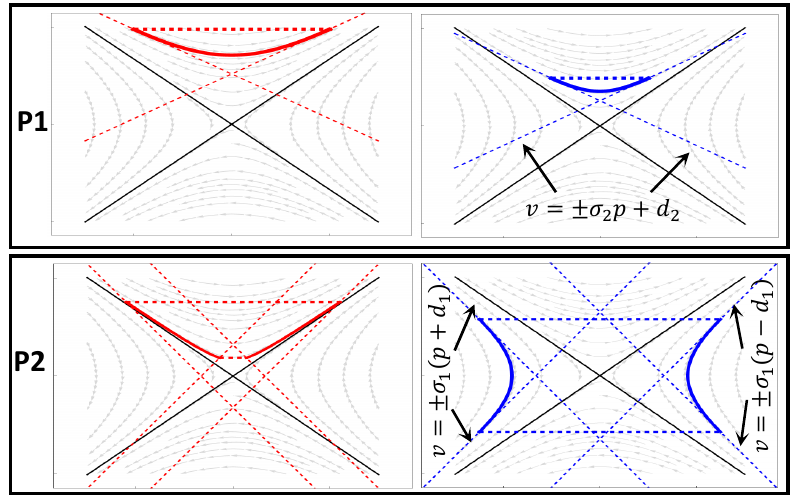}%
      \caption{Equivalent characterization of the periodic orbits. The dashed cross lines are the equivalent orbital lines.}
      \label{fig:equivaChar}
\end{figure}

Similarly, P1 orbital lines can characterize P2 orbits:
\\

\noindent\fbox{%
    \parbox{\linewidth-6pt}{%
        \textbf{Proposition 4.} The extended P1 orbital lines $v = \pm \sigma_{1} (p \pm d_1)$ characterize the P2 orbits: the initial states are on $v = -\sigma_1 (p \pm d_1)$ and the final states are on $v = \sigma_1 (p \pm d_1)$. The corresponding step sizes are as stated in \eqref{eq:stepLength2}.
    }
}
\\

The non-uniqueness of P2 orbits to realize the desired velocity comes from the non-uniqueness of $d_1$. Given a $d_1$, the final states of the P2 orbits can thus be determined. Fig. \ref{fig:equivaChar} illustrates the equivalent characterizations of the orbits in Fig. \ref{fig:LIP_p1} and \ref{fig:LIP_p2}. In the latter, we only use the results from Theorem 1 and 2 to find the desired walking orbits.

\subsubsection{3D Composition}
Full 3D walking can be encoded by two orthogonally composed planar orbits. The desired 3D walking behavior is first described via the desired walking velocities $v^d_{x,y}$ in the sagittal and coronal plane. The orbit that realizes the walking velocity is then identified in each plane. A typical composition is choosing a P1 orbit in the sagittal plane and a P2 orbit in the coronal plane (sP1-cP2). The non-uniqueness of the P2 orbit can prevent foot collisions by selecting step sizes to have opposite signs. This will be the main composition we realize on the 3D robot.

\subsection{Orbit Stabilization}
We now derive the stepping stabilization on the periodic orbits. It can be viewed as generating a controller on $u$ such that the S2S state is controlled to the desired final states in \eqref{eq:desiredP1state} for P1 orbits and in \eqref{eq:desiredP2pos} for P2 orbits. In \cite{xiong2019orbit}, the stabilization was formulated based on the hybrid dynamics, and the proof was on the contraction on the distance between the state and the target state of the orbit. Hence, the ranges of the gain and the optimal gains in terms of contraction rate were derived from the contraction. 
One can also derive the stabilization via the S2S dynamics. It becomes a canonical linear control problem: controlling the state to the desired one based on the linear dynamics. We do not present this approach here. Instead, we directly apply the H-LIP based stepping in \eqref{eq:HLIPstepping} to stabilize the orbits of the H-LIP, which yields the simplest derivation. The stepping stabilization for P1 orbits is:
\begin{equation}
\label{eq:orbitStbP1}
    u = u^* + K(\mathbf{x} - \mathbf{x}^*),
\end{equation}
where $\mathbf{x} = [p^-,v^-]^T$ is the current pre-impact state of the H-LIP, $u^*$ is the step size of the desired P1 orbit, and $\mathbf{x}^* = [p^*, v^*]^T$ is the pre-impact state of the desired P1 orbit. The error state is $\mathbf{e} = \mathbf{x} -  \mathbf{x}^*$, and the error S2S dynamics becomes: 
\begin{equation}
\label{eq:orbitErrorDynamics}
    \mathbf{e}_{k+1} = (A+BK)\mathbf{e}_k.
\end{equation}
This is \eqref{eq:errorDynamicsHLIPstepping} with $w = 0$ since the H-LIP stepping is applied on the H-LIP itself. To drive the H-LIP to its orbit, i.e., $\mathbf{e} \rightarrow 0$, we only need to find $K$ to make $A+BK$ stable. The deadbeat control can be applied. Since the system has two states and one input \rev{and it is controllable (its controllability matrix is of full rank)}, it requires to two steps to make $\mathbf{e} \rightarrow 0$ for all $\mathbf{e} \in \mathbb{R}^2$. The deadbeat gain is calculated from:
$(A+BK_\text{deadbeat})^2 =0,$
which yields, 
\begin{equation}
\label{eq:deadbeatGain}
K_\text{deadbeat} =\begin{bmatrix}
1 &  T_{\text{DSP}} +  \frac{1 } {\lambda}\text{coth}(T_{\text{SSP}} \lambda)
\end{bmatrix}.
\end{equation}
Plugging the deadbeat gain into \eqref{eq:orbitStbP1} yields, 
\begin{align}
\label{eq:deadbeatP1}
    u_k = \textstyle p + p^* + T_{\text{DSP}} v + \frac{1}{\lambda} \text{coth}(T_{\text{SSP}} \lambda) (v- v^*).
\end{align}
This is verified to be equal to the optimal stepping controller in Theorem 2.1 in \cite{xiong2019orbit}, which globally stabilizes the system to the desired P1 orbit with two steps. Similarly, the stepping stabilization for P2 orbits is: 
\begin{equation}
    u_\text{L} = u^*_\text{L} + K(\mathbf{x}_\text{L} - \mathbf{x}^*_\text{L}), 
    \quad u_\text{R} = u^*_\text{R} + K(\mathbf{x}_\text{R} - \mathbf{x}^*_\text{R}), \label{eq:HLIPstablization}
\end{equation}
which yields the same error dynamics in \eqref{eq:orbitErrorDynamics}. When the deadbeat gain in \eqref{eq:deadbeatGain} is chosen, the controller becomes identical to the optimal controller in Theorem 2.2 in \cite{xiong2019orbit}.

This application of the "H-LIP stepping" on the H-LIP should not be confused to that on the robot. On the robot, we initialize the H-LIP state to be identical to the robot, stabilize the H-LIP to the desired behavior \rev{using \eqref{eq:HLIPstablization}}, and then stabilize the robot to the walking of the H-LIP \rev{using \eqref{eq:HLIPstepping}}.

\noindent{\underline{\textit{Comparison to Capture Point:}}} The deadbeat stepping controller on the H-LIP is similar to the capture point controller \cite{pratt2012capturability}. In the capture point controller, the step location is determined by the passive LIP so that the robot can come to a stop, i.e., $v \rightarrow 0$ as $t\rightarrow \infty$. In comparison, the H-LIP with zero velocity is a P1 orbit with $v^* =0$. Additionally, if we assume $T_{\text{SSP}} \rightarrow \infty$ and $ T_{\text{DSP}} \rightarrow 0$, \eqref{eq:deadbeatP1} becomes identical to the instantaneous capture point controller: 
\begin{equation}
\label{eq:LIPtoCapturePoint}
u ={\textstyle  p +  \cancelto{0}{p^*} + \cancelto{0}{T_{\text{DSP}}} v + \frac{1}{\lambda} \cancelto{1}{\text{coth}(T_{\text{SSP}} \lambda)} (v- \cancelto{0}{v^*}) }= p +  \frac{v}{ \lambda}. \nonumber   
\end{equation}
Thus, the capture point controller on the H-LIP is a special case of the deadbeat stepping controller on this model. More importantly, the capture point controller based on the LIP is typically directly applied on the robot; whereas the stepping controller on the H-LIP, e.g. in \eqref{eq:deadbeatP1}, is not directly applied on the robot but is used at the nominal step size $u^\text{H-LIP}$ in \eqref{eq:HLIPstepping}. 


\textbf{Remark 1:} The S2S formulation of H-LIP stepping for its orbit stabilization also enables the use of many linear controllers. For instance, the Linear Quadratic Regulator (LQR) controller \cite{boyd1991linear, xiong2020ral} can be applied to provide the optimal gain $K$ subject to a quadratic cost on the states and inputs. Model predictive controllers (MPC) \cite{camacho2013model} can be easily synthesized on the linear S2S dynamics to directly stabilize the state to the desired one on the periodic orbits. In this paper, we do not present those results or their comparisons. \rev{Instead, we directly realize the deadbeat controller on the H-LIP and then on the robot Cassie to evaluate the H-LIP based approach.}

\section{Application to Bipedal Robot Cassie}
\label{sec:Cassie}

\subsection{The Robot Model} 
\begin{figure}[b]
      \centering
      \includegraphics[width = 0.9\columnwidth]{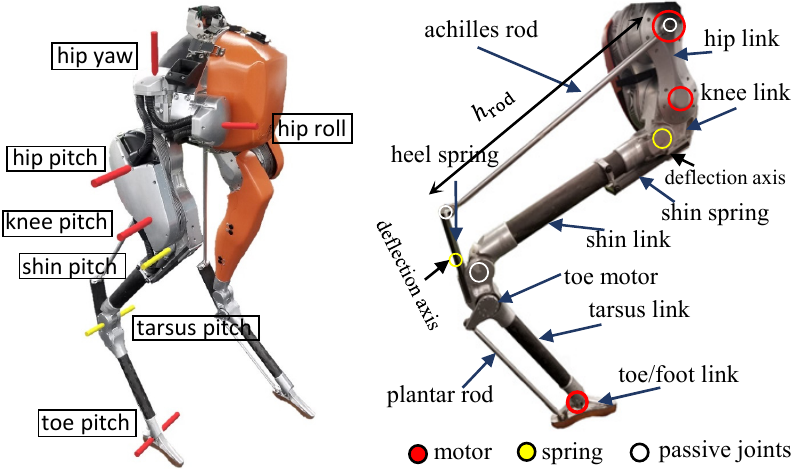}
      \caption{The underactuated bipedal robot Cassie, its joints, and linkages.}
      \label{fig:cassie}
\end{figure}

The robot Cassie (Fig. \ref{fig:cassie}) is a 3D underactuated bipedal robot with compliance. It is designed and built by Agility Robotics \cite{AG} to resemble the SLIP model \cite{geyer2006compliant, rezazadeh2018robot} for locomotion; \rev{the company provided the kinematics and inertia of each linkage of the robot, which are used in the articulated rigid-body model in our study.} Roughly speaking, the robot has a concentrated upper body and light-weight springy legs. 
Here, we describe the mathematical model that best captures its dynamics with certain simplifications. 

Each leg on the robot can be modeled with seven degrees of freedom, including five motor joints and two leaf springs. The two leaf springs can be modeled as two rotational joints with torsional springs \cite{xiong2018bipedal, reher2020inverse}. As shown in Fig. \ref{fig:cassie}, the four-bar linkage on the lower part of the leg transmits a distant motor actuation to the pitch of the foot. The closed-loop linkage on the upper part of the leg can be viewed as to translate the rotation of the knee joint to the tarsus joint, which extends and retracts the foot. Since the push-rods are light-weighted, we neglect their inertia and their associated degrees of freedom (dofs) for simplifications. The foot is then assumed to be directly actuated from the toe motor. The Achilles rod is replaced by a holonomic constraint on the distance between the end-points of the rod. 

We use the floating-base coordinate to describe the configuration of the robot: $q = [q_{\text{pelvis}}, q^\text{L}_{\text{leg}}, q^\text{R}_{\text{leg}}]^T$, where $q_\text{pelvis} = [q^{x,y,z}_\text{pelvis}, q^{rpy}_\text{pelvis}] \in SE(3)$ is the pelvis configuration, and $q^\text{L}_{\text{leg}}, q^\text{R}_{\text{leg}}$ are the configuration of the left and right leg, respectively. $q^{\text{L/R}}_{\text{leg}} = [q^{\text{roll}}_{\text{hip}},q^{\text{yaw}}_{\text{hip}},q^{\text{pitch}}_{\text{hip}},q_{\text{knee}}, q_{\text{shin}}, q_{\text{tarsus}}, q_{\text{heel}}, q_{\text{toe}}]$, where the individual element is the joint angle. 
The motor joints are $q_{\text{motor}} = [q^{\text{roll}}_{\text{hip}},q^{\text{yaw}}_{\text{hip}},q^{\text{pitch}}_{\text{hip}},q_{\text{knee}}, q_{\text{toe}}]^T$, and the spring joints are $q_{\text{spring}} = [ q_{\text{shin}}, q_{\text{heel}}]^T$.
The continuous dynamics are:
\begin{align}
\label{eq:robotEOM}
& M(q) \ddot{q} + C(q, \dot{q}) + G(q) = \rev{B \tau_m + J^T_s \tau_s}  + J_h^T F_h, \\
& J_h \ddot{q} + \dot{J}_h \dot{q} = 0, \label{eq:robotHolonomic} 
\end{align}
where $M(q)$ is the mass matrix, $C(q, \dot{q})$ contains the Coriolis and centrifugal forces, $G(q)$ is the gravitational vector, \rev{$\tau_m$ represents the motor torque vector, $\tau_s$ is the vector of the torsional forces of the spring joints, $J_s$ is the Jacobian of the spring joints, $B$ is the actuation matrix,} $J_h = [J_\text{rod}^T, J_\text{Foot}^T]^T$ represents the Jacobian of the holonomic constraints, and $F_h = [F_\text{rod}^T, F_\text{GRF}^T]^T$ contains the holonomic forces, including the forces on the push-rods and ground reaction forces (GRF). The spring forces are calculated by: 
$
    \tau_{\text{shin/heel}} = \text{K}^s_{\text{shin/heel}}~q_{\text{shin/heel}} + \text{D}^s_{\text{shin/heel}} ~\dot{q}_{\text{shin/heel}},
$
where $\text{K}^s_{\text{shin/heel}}$ and $\text{D}^s_{\text{shin/heel}}$ are the stiffness and damping of the springs. 
The holonomic constraints $h(q)$ include the distance constraints on the Achilles push-rods and the ground contact constraints. In the SSP, the ground contact can be described via 5 holonomic constraints, and the dimension of $h(q)$ is 7. In DSP, the dimension of $h(q)$ becomes 12. Note that, \eqref{eq:robotEOM} and \eqref{eq:robotHolonomic} provide an affine mapping from the input torques to the holonomic forces:
\begin{equation}
\label{eq:holonomicForceFromTorque}
    F_h = A_h \tau_m + b_h.
\end{equation}
The exact expressions of $A_h, b_h$ are omitted here. 

\noindent{\underline{\textit{Hybrid Walking Model:}}} The walking of Cassie is modeled as a two-domain hybrid system. Due to the compliance in the legs, the transition from the current SSP to the next SSP is not likely to instantaneously happen right after the impact event. In other words, the DSP typically exists in walking; \rev{hypothetically if there is no compliance considered, the walking can be modeled as a one-domain hybrid system with a trivial DSP.} The impact between the swing foot and the ground is assumed to be plastic \cite{grizzle2014models}. 
Note that the foot is small and its rotation is not actuated in the lateral direction. Thus, the walking in the coronal plane is underactuated at the foot. The toe actuation on the stance foot is virtually removed by setting the torque to 0 to render foot-underactuation in the sagittal plane as well.

\subsection{H-LIP based \rev{Direct} Gait Design and Stepping on Cassie}
\begin{figure}[b]
    \centering
    \includegraphics[width = .9\columnwidth]{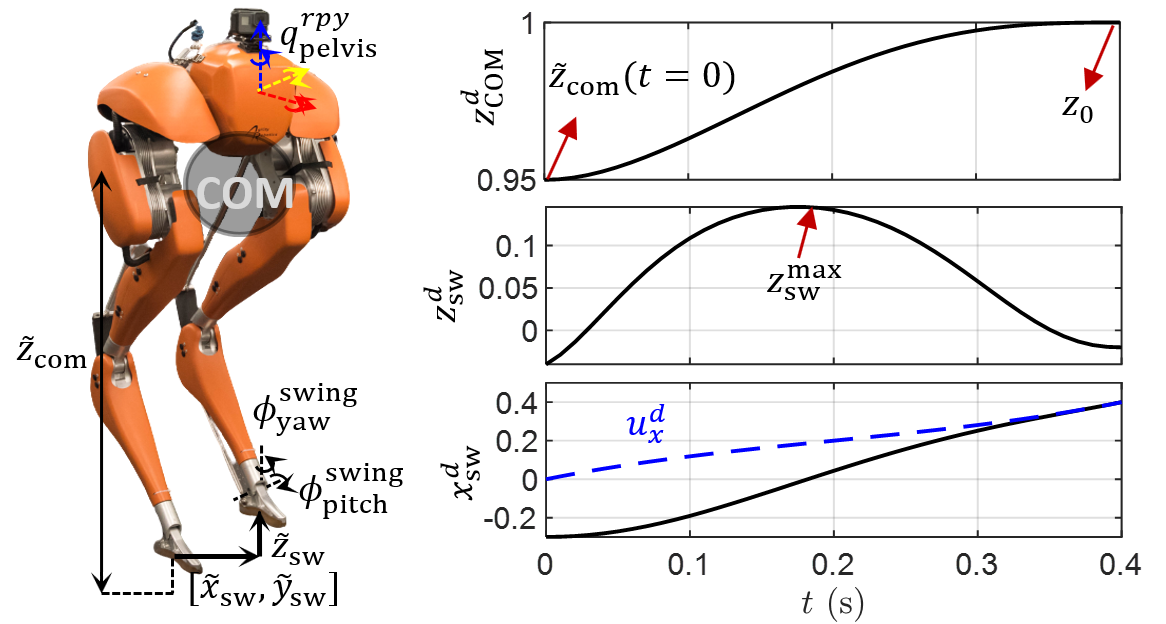} 
    \caption{Illustrations of the definition and desired trajectories of the output.}
    \label{fig:cassieOutputs}
\end{figure}

Now we apply the H-LIP based approach on Cassie. The output is designed to satisfy the requirement of the H-LIP based gait synthesis: the vertical center of mass (COM) position $z_{\text{COM}}$ (w.r.t. the stance foot) should be (approximately) constant, and the vertical position of the swing foot $z_{\text{sw}}$ is constructed to periodically lift-off and strike the ground. The horizontal position of the swing foot $\{x,y\}_{\text{sw}}$ (w.r.t. the stance foot) is controlled to achieve the desired step size $u_{x,y}^d$ from the H-LIP based stepping. Additionally, the orientation of the pelvis $q^{rpy}_\text{pelvis}$ and the swing foot $\phi^\text{sw}_{py}$ should be controlled to fully constrain the walking. The output in SSP (illustrated in Fig. \ref{fig:cassieOutputs}) is then defined as: 
\begin{equation}
\label{eq:cassieOutputSSP}
    \mathcal{Y} =  \begin{bmatrix}
     z_\text{COM} \\
    \{ x,y,z\}_\text{sw} \\
    q^{rpy}_\text{pelvis} \\
     \phi^\text{sw}_{py} 
    \end{bmatrix}-
     \begin{bmatrix}
     z^d_\text{COM}  \\
  \{ x,y,z\}^d_\text{sw} \\
      q^{{rpy}^d}_\text{pelvis} \\
     {\phi^\text{sw}_{py}}^d
    \end{bmatrix}.
\end{equation}

\subsubsection{Accommodations for Compliance}
The output definition in \eqref{eq:cassieOutputSSP} is sufficient for robots without evident compliant elements. However, the passive compliance on Cassie creates challenges on precise control on the vertical COM and swing foot positions. If the output contains the compliant degrees of freedom, the spring can create undesired resonance, which then destabilizes the output, especially in the vertical direction. Thus, accommodations have to be made for the compliance. 

In \cite{xiong2019orbit}, the uncompressed leg length (i.e. the leg length with zero spring deflections) was used as the output to indirectly control the vertical position of the COM and the swing foot. Here we select the uncompressed vertical COM and swing foot positions as the approximation to the actual ones. By definition, the vertical position of the COM w.r.t. the stance foot is a function of $\{q^{rpy}_\text{pelvis}, q_\text{motor},q_\text{tarsus}, q_\text{spring}\}$. The COM height with uncompressed springs is: $$\tilde{z}_\text{COM} = z_\text{COM}(q^{rpy}_\text{pelvis}, q_\text{motor}, q_\text{tarsus} \rightarrow q^\text{rigid}_\text{tarsus}, q_\text{spring} \rightarrow 0).$$ $q^\text{rigid}_\text{tarsus}$ is the uncompressed tarsus angle under the holonomic constraint of the push-rod: 
\begin{equation}
q^\text{rigid}_\text{tarsus} = \texttt{Root}(h_\text{rod}(q_\text{knee}, q_\text{shin} \rightarrow 0,  q_\text{heel} \rightarrow 0, q_\text{tarsus}) = 0), \nonumber
\end{equation} 
which is solved via Newton-Raphson method. $z_\text{COM}$ in \eqref{eq:cassieOutputSSP} is thus approximated by $\tilde{z}_\text{COM}$. Similarly, the position of the swing foot w.r.t. the stance foot are approximated in the same way by $\{\tilde{x},\tilde{y},\tilde{z}\}_\text{sw}$. Since the springs on the stance leg are expected to oscillate less, we only set the springs on the swing legs to 0 in $\{\tilde{x},\tilde{y},\tilde{z}\}_\text{sw}$ for better approximations. 

\subsubsection{Desired Output Trajectories} The desired orientations of the pelvis and swing foot are chosen to be constant. The rest of the desired trajectories are designed with Bézier polynomials to satisfy the requirements of the H-LIP based approach. The exact Bézier coefficients are listed in the Appendix. 

First, the desired step sizes in the sagittal and coronal plane are constantly decided from the H-LIP based stepping in \eqref{eq:HLIPstepping}: 
\begin{align} 
\label{eq:robotDesiredStepSize}
u_{x/y}^d &= u_{x/y}^\text{H-LIP} + K(\mathbf{x/y}^\rev{R} - \mathbf{x/y}^\text{H-LIP}),
\end{align}
where $\mathbf{x/y}^\rev{R}$ is the horizontal COM state of the robot in the sagittal or coronal plane. The desired horizontal trajectories of the swing foot are designed as:
\begin{align}
\label{eq:desiredSwingX}
    {x/y}_\text{sw}^d &= (1- b_h(t)) {x/y}^+_\text{sw} + b_h(t) u_{x/y}^d,
\end{align}
where ${x/y}^+_\text{sw}$ is the horizontal position of the swing foot w.r.t. the stance foot in the beginning of the current SSP. $b_h(t)$ is a Bézier polynomial that transits from 0 ($t = 0$) to 1 ($t = T_\text{SSP}$), where the clock of the gait $t$ is reset to 0 after each step. 

The vertical COM position should be controlled to $z_0$, which is also the constant height of the H-LIP. At swapping support legs, $\tilde{z}_\text{COM}$ has a small discrete jump. The desired trajectory of the vertical COM position is then constructed as: 
\begin{equation}
z^d_\text{COM} = (1-b_h(t)) \tilde{z}^+_\text{COM} + b_h(t) z_0,
\end{equation}
where $\tilde{z}^+_\text{COM}$ is the uncompressed COM height in the beginning of the SSP. Lastly, the vertical position of the swing foot $z^d_\text{sw}(t)$ is constructed as: 
\begin{equation}
z_\text{sw}^d(t) = b_v(t, z^\text{max}_\text{sw}, z^\text{neg}_\text{sw}),
\end{equation}
where $b_v$ is another Bézier polynomial to create lift-off and touch-down behaviors. It is designed to transit from 0 ($t = 0$) to $z^\text{max}_\text{sw}$ (e.g., $t = \frac{T_\text{SSP}}{2}$) and back to $z^\text{neg}_\text{sw}$ ($t = T$). $z^\text{max}_\text{sw}$ is a constant to determine the foot-ground clearance, and $z^\text{neg}_\text{sw}$ is a small negative value to ensure foot-strike at the end. \rev{Note that the same desired output trajectories can be directly applied to robots without compliance.}

\subsubsection{Desired DSP Output} In DSP, two feet contact the ground at all times. With more holonomic constraints on the system, the dimension of the outputs decreases. Instead of re-formulating a different set of DSP outputs, we directly use the SSP outputs and set the desired values of the outputs on the swing foot to be the actual ones (including the horizontal positions and orientation), which preserves the holonomic constraints in the DSP and also simplifies the gait design.  

\noindent{\underline{\textbf{Versatility:}}} The output construction directly allows the COM height $z_0$, step frequency (inverse of the walking period $\frac{1}{T}$) and the swing foot clearance $z^\text{max}_\text{sw}$ to be individually chosen. 
Different combinations of the parameters render different walking behaviors. The desired walking velocity in each plane is individually stabilized via the H-LIP stepping, which is independent of the chosen gait parameters. Additionally, for P2 orbits on the robot, there are infinite orbits for realizing the same desired walking velocity. The combination of the gait parameters and orbit selections renders versatile walking behaviors on the robot. 

\textbf{Remark 2:} In \cite{xiong2019orbit}, a stepping-in-place gait was optimized on the aSLIP and its periodic trajectories of the leg length were then applied on Cassie. Different walking behaviors were realized via perturbing the stepping-in-place gait by changing the step size based on the H-LIP. The periodic leg length trajectories \rev{\textit{indirectly}} realized the lift-off and touch-down behaviors on the swing foot and rendered an approximately constant vertical COM height. Here, the output is constructed in a more direct and general fashion, \rev{which creates a better approximation from the H-LIP to the robot. The same gait design also works on robots without compliance. Moreover, the use of the aSLIP is not necessary, and solving non-convex trajectory optimization is eliminated. Finally, the desired output trajectories are directly constructed in closed-form, where the gait parameters such as step frequency, vertical COM height, and swing foot clearance can be directly changed continuously.} 

\subsection{Joint-Level Optimization-based Controller}
Nonlinear controllers can be applied to drive the output $\mathcal{Y}$ in \eqref{eq:cassieOutputSSP} to zero. In particular, we consider using Quadratic Program (QP) based controllers \cite{wensing2013generation, ames2014rapidly} for stabilization, where the contact constraints and torque limits can be included as the inequality constraints in the QP. When the foot contacts the ground, the resultant ground reaction forces should satisfy the friction cone constraints and the non-negativity constraints on the normal forces. This can be encoded as:
  $ \mathcal{A} F_\text{GRF} \leq 0,$
where $\mathcal{A}$ is a constant matrix. The motors can only provide certain amount of torques at certain speeds. Thus, we set:
 $  \tau_\text{lb}(\dot{q}) \leq \tau_m(\dot{q}) \leq \tau_\text{ub}(\dot{q})$,
where $\tau_\text{lb/ub}$ is the lower or upper bounds on the motor torques.  

The control objective is to drive the output $\mathcal{Y} \rightarrow 0$. Here we illustrate two prominent approaches for realization: one is in the task space control (TSC) formulation \cite{wensing2013generation} through minimizing the difference between the actual acceleration and a desired acceleration, which yields stable linear dynamics on the output; the other one is in the control Lyapunov function (CLF) formulation \cite{ames2014rapidly} via an inequality condition on the derivative of the Lyapunov function of the output to yield exponential convergence. 

For the TSC, the desired acceleration $\ddot{\mathcal{Y}}^d$ is chosen as: 
\begin{align}
\label{eq:outputLinearTSC}
\ddot{\mathcal{Y}}^d = - \rev{\mathcal{K}_p} \mathcal{Y} - \rev{\mathcal{K}_d }\dot{\mathcal{Y}}
\end{align}
with \rev{$\mathcal{K}_p, \mathcal{K}_d$ being the feedback PD gains}. An optimization problem is formulated to minimize $||\ddot{\mathcal{Y}} - \ddot{\mathcal{Y}}^d||^2$ subject to the physical constraints and the robot dynamics. Then the actual output dynamics evolves closely to the desired linear stable dynamics in \eqref{eq:outputLinearTSC}, which thus realizes the control objective. Since the acceleration $\ddot{\mathcal{Y}}$ is affine w.r.t. the input torque, the optimization problem is a quadratic program (QP). 

For the CLF, a quadratic Lyapunov function $V(\mathcal{Y}, \dot{\mathcal{Y}})$ is constructed on the output $\mathcal{Y}$ and $\dot{\mathcal{Y}}$, thus $V(\mathcal{Y}, \dot{\mathcal{Y}})\rightarrow0$ if and only if $[\mathcal{Y}, \dot{\mathcal{Y}}]\rightarrow0$. The convergence condition of $\mathcal{Y}$ is enforced via the derivative of $V$, i.e., 
\begin{align}
\label{eq:dotVcondi}
   \dot{V} \leq - \gamma V,
\end{align}
with $\gamma > 0$, which yields $V$ (and thus $\mathcal{Y}$) to decrease at least at an exponential rate. As $\dot{V}$ is affine w.r.t. the input torque, a QP can be formulated on minimizing the norm of the input torque subject to the inequality constraint in \eqref{eq:dotVcondi} and additional physical constraints. The two QP based controllers are summarized as follows.
\begin{table}[h]
    \centering
    \begin{tabular}{|c|c|} 
    \hline
    TSC-QP  & CLF-QP \\
    \hline
    $[\tau_m^*, \sim]  = \underset{\tau_m, F_h, \ddot{q}}{\text{argmin}} \ ||\ddot{\mathcal{Y}} - \ddot{\mathcal{Y}}^d||^2$ 
    & 
    $[\tau_m^*, \sim]  = \underset{\tau_m, F_h}{\text{argmin}} \  \tau_m^T \tau_m$ \\
      $\begin{aligned}
       \text{s.t.}  \ & \text{Eq.} \eqref{eq:robotEOM}, \eqref{eq:robotHolonomic},\mathcal{A} F_\text{GRF} \leq 0  \\
         & \tau_\text{lb}(\dot{q}) \leq \tau_m(\dot{q}) \leq \tau_\text{ub}(\dot{q}) 
      \end{aligned}$  
      &  
      $\begin{aligned}
     \quad \text{s.t.}\  &   \text{Eq.} \eqref{eq:holonomicForceFromTorque},\eqref{eq:dotVcondi},  \mathcal{A} F_\text{GRF} \leq 0 \\
         & \tau_\text{lb}(\dot{q}) \leq \tau_m(\dot{q}) \leq \tau_\text{ub}(\dot{q})
      \end{aligned}$    \\
      \hline
    \end{tabular}
\end{table}

\textbf{Remark 3:} We do not intend to compare the two controllers or propose any other variants of the QP-based controllers in this paper. They are merely used as tools to stabilize the output to demonstrate the H-LIP based approach. With proper gain-tuning, both controllers can perform equivalently. 

\section{Simulation Evaluation}
\label{sec:Sim}
\begin{figure*}[t]
    \centering
    \includegraphics[width = .9\textwidth]{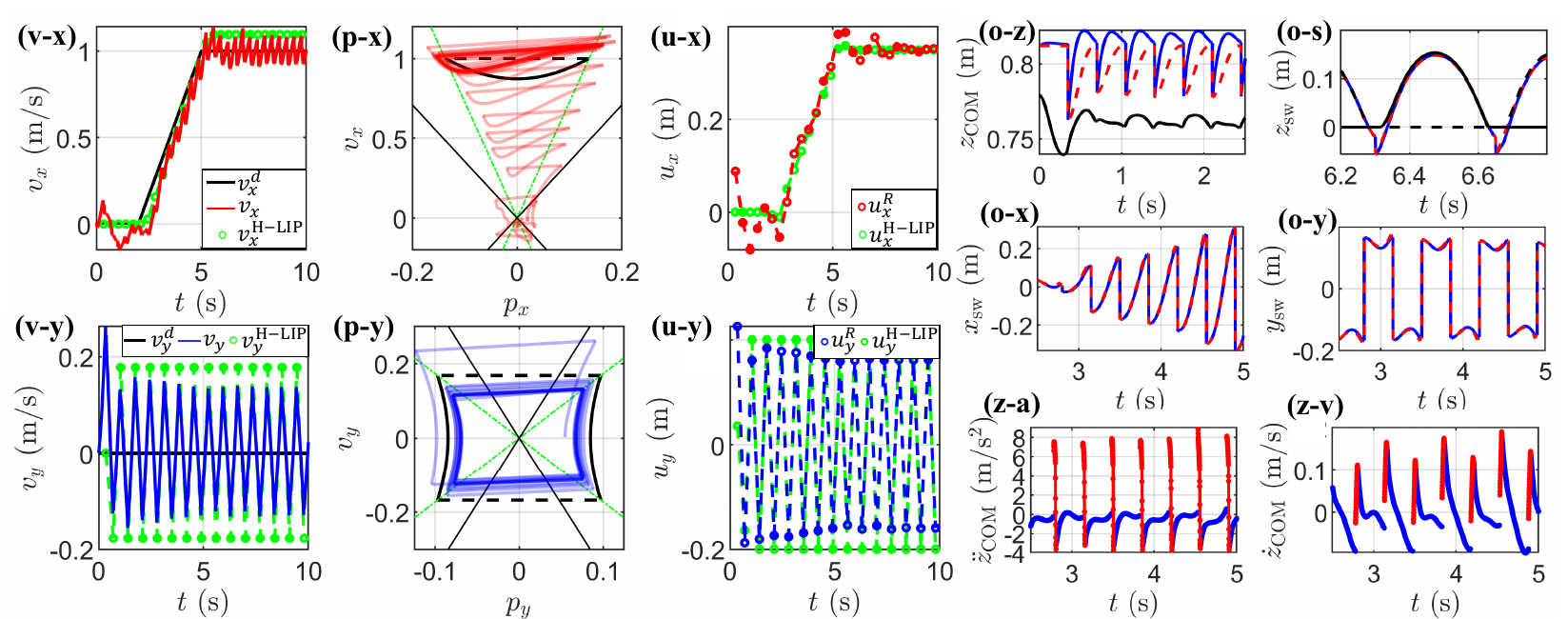}
        \caption{Simulation results on a forward walking with $v^t_{x,y} = [1,0]$m/s, ${u^y_\text{L}}^* = -0.2$m: the trajectories of the horizontal velocities of the COM (red and blue lines) in the sagittal plane (v-x) and the coronal plane (v-y) compared with the desired velocities $v^d_{x,y}(t)$ (black lines) and the corresponding velocities of the H-LIP (green circles); the phase trajectories of the horizontal states of the COM in the sagittal plane (p-x) and the coronal plane (p-y) compared with the H-LIP orbits (black) at the target velocities; comparisons of the step sizes (u-x, u-y) between the robot (red circles in the sagittal plane and blue circles in the coronal plane) and the H-LIP (green circles). (o) Output tracking with the red dashed lines indicating the desired output trajectories and the blue continuous lines indicating the actual one: (o-z) the vertical COM trajectory $\tilde{z}_\text{COM}$ (the black line is the actual vertical COM position of the robot $z_\text{COM}$); (o-s) the vertical swing foot trajectory $\tilde{z}_\text{sw}$ (the black lines are $z_\text{sw}$); the horizontal trajectories of the swing foot in the sagittal plane (o-x) and the coronal plane (o-y). \rev{(z) Actual vertical COM acceleration (z-a) and velocity (z-v) where the blue and red are the SSP and DSP trajectories, respectively.} }
    \label{fig:simForwardRobotVsLIP}
\end{figure*}

We now realize and evaluate the approach on Cassie in simulation. The simulation environment allows thorough evaluations on the robot model at the stage before hardware realization. In the simulation, we have full access to all the states of the system. Thus, the information of contacts and the horizontal velocity of the robot are exactly known, which provides a rigorous analysis of the proposed approach.

\noindent{\textbf{Setup}:}
The robot starts from a static standing configuration. The dynamics of the robot are numerically integrated using the ODE 45 function in Matlab with event-based functions for triggering domain switching. Target final velocities $v^t_x$ and $v^t_y$ are given with the goal of controlling the robot to realize these target velocities. We first select an orbit composition and then design continuous desired velocity profiles $v^d_{x/y}(t)$ (by piecewise linear trajectories for simplicity) to reach the target velocities. For P2 orbits, the desired step size should be specified. The desired output trajectories are constructed via the H-LIP based gait synthesis and stepping. The gait parameters such as the swing foot clearance and step frequency are specified in the beginning. The low-level optimization-based controller is solved at 1kHz using qpOASES \cite{Ferreau2014}. The video of the simulation results can be seen in \cite{video:sim}.

\subsubsection{Forward Walking}
 We first present forward walking as the basic realization of the proposed H-LIP based approach. The orbit composition is chosen as having a P1 orbit in its sagittal plane and a P2 orbit in its coronal plane. The velocities are chosen to be $v^t_x = 1$m/s, and $v^t_y = 0$m/s, thus the robot only progresses in its sagittal plane. We choose $T = 0.35$s, $z^\text{max}_\text{sw} = 0.15$m,  $z^\text{neg}_\text{sw} = -0.02$m, and the orbit-determining step width of the P2 orbit is $u^{y*}_\text{L} = -0.2$m. The desired walking velocity $v^d_x(t)$ is chosen from 0 to ramp up to 1m/s within 3s.  Fig. \ref{fig:simForwardRobotVsLIP} plots the walking trajectories via the H-LIP based approach. \rev{The H-LIPs are controlled to the desired walking and then the robot is controlled to the walking of the H-LIP.} The output trajectories are well-tracked via the optimization-based controller, \rev{and the resulting vertical COM has small velocities and accelerations.} The horizontal COM states of the robot converge with negligible errors to the desired H-LIP states that realize the target velocities in each plane.  
 
\begin{figure}[!t]
    \centering
    \includegraphics[width = 1\columnwidth]{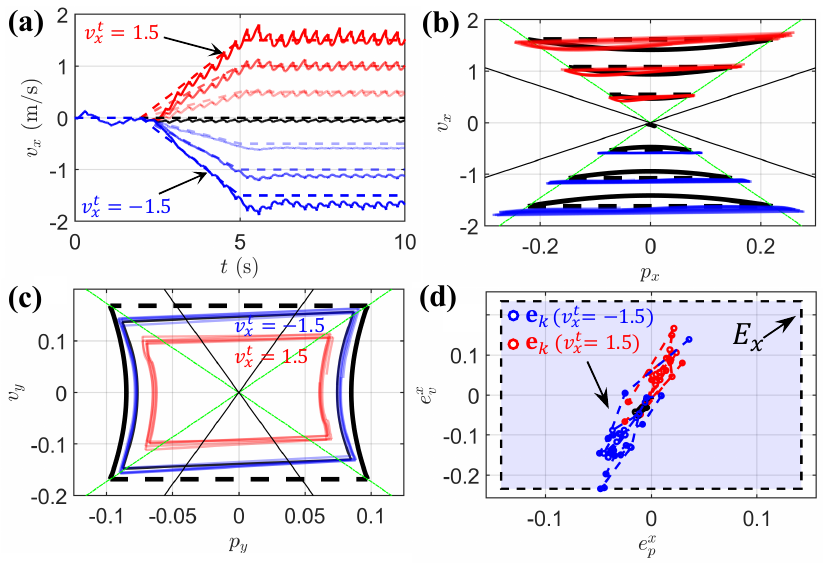} 
    \caption{Comparison on forward walking with different target velocities $v^t_x$: (a) forward velocities of the COM (continuous lines) compared with the desired velocity profiles $v^d_x(t)$ (dashed lines); (b) the converged orbits (red and blue lines) of the sagittal COM states compared with the desired target orbits of the H-LIP (black lines). (c) the converged orbits (the red is with $v^t_x= 1.5$ and the blue is with $v^t_x = -1.5$) of the coronal COM states compared with the target orbit of the H-LIP (black); (d) the error state trajectories (circles) and the error invariant set $E_x$ (the blue transparent box) in the sagittal plane.}
    \label{fig:simForwardCompare}
\end{figure}

 Then we change the target velocity in the sagittal plane from -1.5m/s to 1.5m/s with a 0.5m/s increment. Fig. \ref{fig:simForwardCompare} shows the results. For clarity, in the phase portraits, we only plot the steady walking behavior where the desired walking velocity becomes constant (after 5s). We also demonstrate that the evolution of the error states is within the error invariant set. The error states in each plane can be directly calculated from the pre-impact states of the robot and the desired states of the H-LIPs. \rev{To calculate the error invariant set, we first numerically calculate the dynamics error $w$ in \eqref{eq:RobotS2Sapprox} from the evolution of the horizontal COM states in each realized walking behavior: $w_k = \mathbf{x}^R_{k+1} - A\mathbf{x}^R_{k} - B u^R_k$}. Since $W$ cannot be calculated analytically, we use all $w$ to construct a polytope to numerically approximate $W$ in each plane. As $K$ is chosen from the deadbeat controller, i.e., $(A+BK)^2 = 0$, the invariant set 
$     E = (A+BK) W \oplus W
$. The set operation is calculated using the MPT \cite{MPT3} toolbox. Fig. \ref{fig:simForwardCompare} (d) shows that the error states are indeed inside the error invariant sets. 

\subsubsection{Lateral and Diagonal Walking} 
\begin{figure}[!t]
      \centering
      \includegraphics[width = 1\columnwidth]{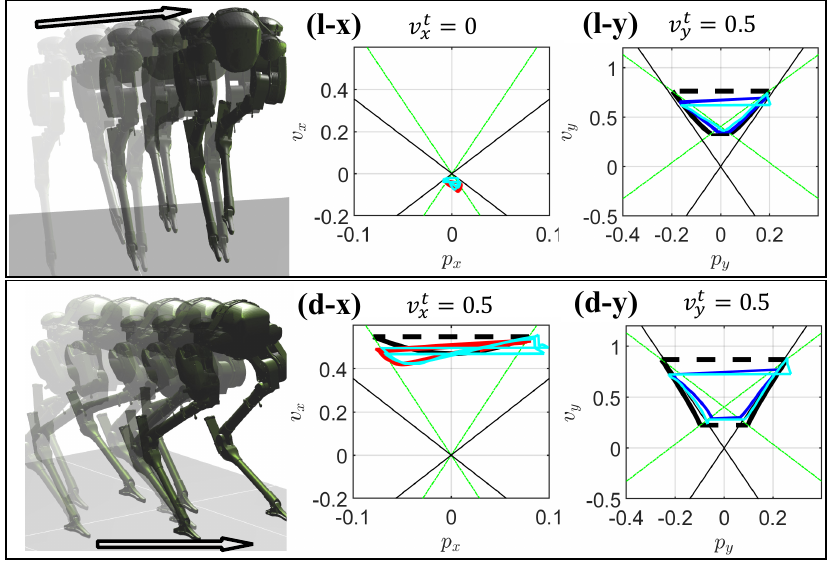} 
      \caption{\rev{Lateral (${u^y_\text{L}}^* = -0.08$m) and diagonal (${u^y_\text{L}}^* = -0.3$m) walking with their converged orbits (the red in the sagittal plane (l-x, d-x) and the blue in the coronal plane (l-y, d-y)). The cyan are the converged orbits of walking on Cassie without compliance springs.} }
      \label{fig:sim_directional}
\end{figure}

The approach can also realize walking to different directions by selecting different desired velocities in each plane of the robot. Here we present walking in the lateral and diagonal directions. Fig. \ref{fig:sim_directional} illustrates the converged walking behaviors with different choices of the target velocities. The gait parameters are identical to the previous examples. By tracking the desired velocity in each plane, the robot walks in the desired direction. The converged orbits of the horizontal COM states are also relatively close to the desired orbits of the H-LIP in both cases. Moreover, by selecting different desired step width $u^{y*}_L$, different P2 orbits are realized in the coronal plane with the same desired velocity $v_y^d$ (Fig. \ref{fig:sim_directional} (l-y) and (d-y)).

\subsubsection{Variable Orbit Compositions} 
\begin{figure}[t]
      \centering
      \includegraphics[width = 0.65\columnwidth]{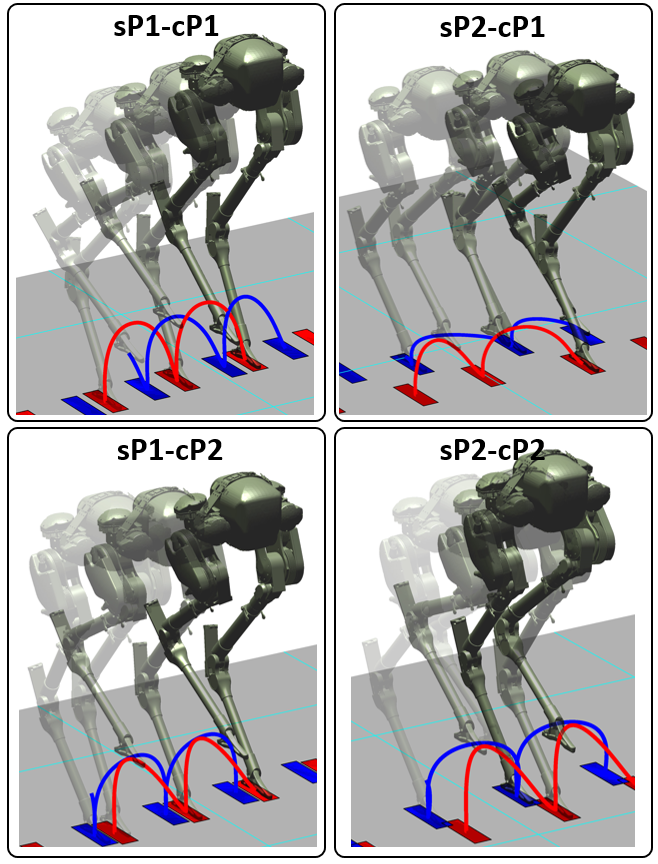}
      \caption{Simulated walking via different orbit compositions with the same target velocity ($v^t_{x,y} = [0, 0.5]$m/s). The trajectories of the swing foot are indicated by the red (left foot) and the blue (right foot) lines, with the rectangles indicating the contact locations.} 
      \label{fig:simOrbitComposition}
\end{figure}

The two types of orbits of the H-LIP provide four kinds of orbit compositions in 3D. If the kinematic constraints are neglected, all four types of orbit compositions can be realized to achieve the same desired walking velocity. Fig. \ref{fig:simOrbitComposition} illustrate the four realizations to achieve the lateral walking with $v^t_y = 0.5$m/s. Each realization is abbreviated by the type of orbit in each plane, e.g., sP1-cP1 indicates both P1 orbits are selected in the sagittal and coronal plane. For certain compositions, kinematic collisions can happen between the legs. E.g., the sP1-cP1 gait with $v^t_{x,y} = 0$ clearly creates foot overlaps. The complex leg design on Cassie further increases the possibilities of kinematic collisions between the legs. Although it is still possible to realize those compositions with certain orbits, we only focus on the realization of the sP1-cP2 orbits on the hardware.

\rev{\textbf{Remark 4:} As an additional verification, we also evaluate the controller on a "virtually rigidified" Cassie without compliance in simulation by fixing the spring joints by holonomic constraints (letting $J_h = [J^T_s, J^T_\text{rod}, J^T_\text{Foot}]^T$ and $F_h = [\tau_s^T, F_\text{rod}^T, F^T_\text{GRF}]^T$ in \eqref{eq:robotHolonomic}). The walking is then modeled as a one-domain system ($T_\text{SSP} = T, T_\text{DSP} = 0$). The gait is synthesized under the same desired trajectories only without the DSP parts; all parameters are kept exactly the same as these in the compliant walking. Similar results are produced in terms of horizontal COM state tracking (see Fig \ref{fig:sim_directional}). After the faithful evaluation in simulation, we next focus on presenting the hardware realization.}

\section{Experiment Evaluation on Cassie} 
\label{sec:exp}
\subsection{Control Realization on Hardware}
Unlike in simulation, the robot state is no longer completely and exactly known on the hardware; \rev{the on-board sensors are the inertia measurement unit (IMU) on the pelvis and the joint encoders in the primary kinematic chain of the leg.} Moreover, the computation capacity of the on-board computer has to be taken into consideration. Therefore, we first address the following problems on the hardware.

\subsubsection{Contact Detection} The robot Cassie is not equipped with contact sensors to detect foot-ground contact, \rev{which makes it challenging for real-time estimation and control.} It is possible to calculate the deflection of the spring joints and set a threshold for contact detection. However, the springs can still have non-trivial deflections in the swing phase due to the inertia forces in the leg. Therefore, the threshold has to be set large enough to avoid false detection of contact. However, this can cause significant late-detection of impacts and early-detection of lift-offs. Instead, we use the measured torque from the input current (similar to the proprioceptive sensing \cite{wensingProprioceptive}) along with the spring deflections to \rev{\textit{approximate} rather than calculate} the contact forces at the feet. A threshold is then set on the magnitude of the forces to detect contact. 
\begin{figure}[t]
    \centering
    \includegraphics[width = 1\columnwidth]{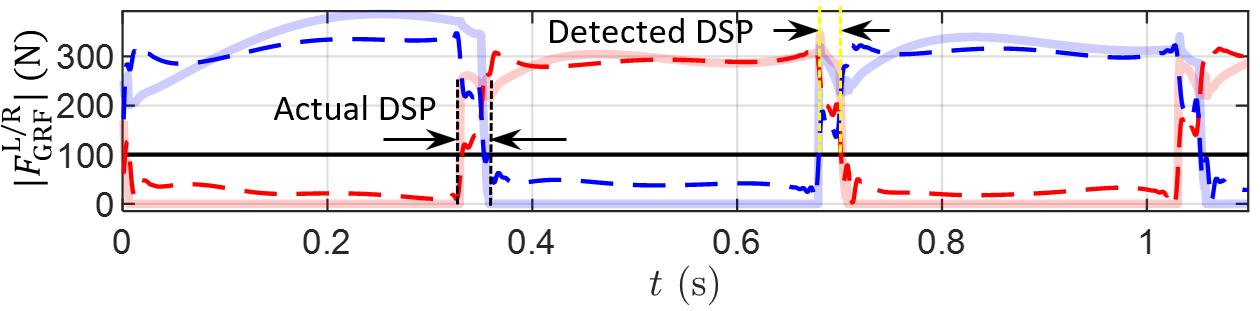}
    \caption{Contact detection via the GRF \rev{approximation}: The transparent red and blue lines are the norm of the actual GRF on the robot in simulation, the dashed red and blue lines are the \rev{approximated} GRF for contact detection. The detected DSP is close to the actual DSP in simulation.}
    \label{fig:expGRFcal}
\end{figure}

\rev{One can also think about using \eqref{eq:robotEOM} to approximate the external forces in a precise manner; however only $\ddot{q}^{x,y,z}_\text{pelvis}$ in $\ddot{q}$ is measured by the IMU and the uncertainty in the inertia modeling can further introduce expected dynamics error in reality.} Since the vertical COM of the robot is controlled approximately constant, we neglect the dynamics contribution to the contact forces. The EOM in \eqref{eq:robotEOM} becomes:
\begin{equation}
\label{eq:staticEOM}
G(q) = \rev{ B \tau_m + J_s^T\tau_s } + J^T_h F_h,
\end{equation} 
where $F_h = [ F_\text{rod}^T, F_\text{GRF}^{\text{L}^T}, F_\text{GRF}^{\text{R}^T}]^T$. $\tau_m$ is measured from the motor current and $\tau_s$ is calculated from the spring deflections. This equation is invariant w.r.t. the pelvis position $q^{x,y,z}_\text{pelvis}$, which is not known. Thus, we set $q^{x,y,z}_\text{pelvis}$ to 0. The rest of $q$ are measured via the IMU, joint encoders and leg kinematics. $F_h$ can be directly solved via the pseudo-inverse of $J^T_h$:
$   F_h = \texttt{pinv}(J^T_h)(G(q) - B \tau_m - J_s^T \tau_s).
$ 
The calculated $F^\text{L/R}_\text{GRF}$ are then low-pass filtered with a cutoff frequency of 100Hz. A threshold is then set on the norm of $F^\text{L/R}_\text{GRF}$ to determine if the foot is in contact with the ground. Fig. \ref{fig:expGRFcal} shows the contact detection via the \rev{approximated GRFs} compared with the actual GRFs in a simulated walking. \rev{Since the dynamics effect is neglected, the magnitudes of the GRFs are not accurate; however, by setting a reasonable threshold (100N), precise contact detections can be achieved during normal walking. }

\subsubsection{H-LIP based Velocity Approximation}
The transitional position and velocity of the floating-base $q^{x,y,z}_{\text{pelvis}}$ and $\dot{q}^{x,y,z}_{\text{pelvis}}$ can not be directly measured. $\dot{q}^{x,y,z}_{\text{pelvis}}$ is required for calculating the COM velocity for realizing the walking. We implemented the extended Kalman filter (EKF) in \cite{bloesch2013state} for state estimation by \rev{fusing the IMU and leg kinematics based on the detected contact}. This state estimation required \rev{non-trivial} computation (a similar estimation scheme in \cite{hartley2018contact} has to be implemented on a secondary computer on the robot). \rev{Moreover, when there is no contact detected (e.g. when walking on unstable terrain), depending on IMU itself quickly induces large velocity estimation errors.} Additionally, the magnetometer drift inside the IMU also creates errors on the estimated velocities under certain circumstances. 
Due to those concerns \rev{and practical hardware limitations}, we instead approximate the COM velocity based on the H-LIP dynamics in the SSP. 

We use the walking in the sagittal plane to illustrate the approximation. Let $p_0$ and $v_0$ be the horizontal position and velocity of the COM in the beginning of the SSP. The dynamics of the horizontal COM in the SSP can be approximated by the SSP dynamics of the H-LIP. Thus the current COM state of the robot $[p_t, v_t]$ in the SSP can be approximated by:
\begin{equation}
\begin{bmatrix}
 p_t, v_t 
\end{bmatrix}^T \approx   e^{A_{\text{SSP}} t} \begin{bmatrix}
 p_0, v_0
\end{bmatrix}^T,
\end{equation}
where $A_\text{SSP}$ is defined in \eqref{eq:LIPsspDynamics}. Let $A_t:= e^{A_\text{SSP}t}$. Given the measured positions $p_0$ and $p_t$ and the current time \rev{$t>0$} from the beginning of the SSP, the velocity approximations are: \begin{equation}
\label{eq:LIPbasedVel}
    \begin{bmatrix}
     \tilde{v}_0 \\ \tilde{v}_t
    \end{bmatrix} = \begin{bmatrix}
     -\nicefrac{A_t^{(1,1)}}{A_t^{(1,2)}} & \nicefrac{1}{A_t^{(1,2)}} \\
   {\scriptstyle A_t^{(2,1)} }- \nicefrac{A_t^{(1,1)} A_t^{(2,2)}}{A_t^{(1,2)}}  & \nicefrac{A_t^{(2,2)}}{A_t^{(1,2)}}
     \end{bmatrix} \begin{bmatrix}
      p_0 \\ p_t
     \end{bmatrix}
\end{equation}
where the superscripts indicate the elements of the matrix $A_t$. Thus the continuous velocity approximation $\tilde{v}_t$ is obtained. The prediction of the pre-impact velocity $\tilde{v}_{t = T_\text{SSP}}$ can also be continuously approximated by the H-LIP dynamics in the SSP based on the current state $[p_t, \tilde{v}_t]^T$ and the time-to-impact $T_\text{SSP} -t$. The velocity approximation is solely based on the position of the COM w.r.t. the stance foot, which only uses joint encoders and orientation readings of the IMU and thus is robust to sensor noises. Moreover, we show that this approximation is valid for applying the H-LIP based stepping, only with generating a different error invariant set.

Let $\tilde{v}^-$ be the approximated velocity of the COM of the robot at the pre-impact. $\tilde{v}^- = \tilde{v}_{t = T_{\text{SSP}}}$ is calculated from \eqref{eq:LIPbasedVel}. Let $\tilde{\mathbf{x}}^\rev{R}  = [p^-,\tilde{v}^-]^T$ represent the approximated COM state at the pre-impact. Assuming the COM position of the robot is measured with a negligible error, $\tilde{\mathbf{x}}^\rev{R}  - \mathbf{x}^\rev{R}  = [0, \tilde{v}^- - v^{-}]^T: = \delta \mathbf{x}$, where $v^{-}$ is the actual COM velocity of the robot at pre-impact event. Note that $\delta \mathbf{x}$ is bounded: the velocity error $\tilde{v}^- - v^{-}$ is the integration of the dynamics difference between the H-LIP and the robot in the SSP. The approximated state is used in the H-LIP based stepping, i.e., $u^\rev{R} = u^{\text{H-LIP}} + K (\tilde{\mathbf{x}}^\rev{R}  - \mathbf{x}^{\text{H-LIP}})$. Therefore, the error S2S dynamics becomes: 
\begin{align}
    \mathbf{e}_{k+1} 
           = (A + BK) \mathbf{e}_k +  \underset{\tilde{w}_k}{\underbrace{w_k + B K \delta \mathbf{x}_k}}. \label{eq:errorDynUnderVelApprox}
\end{align}
$\tilde{w} \in \tilde{W}$ is bounded since $w$ and $\delta \mathbf{x}$ are both bounded. This consequently creates a new error invariant set $\tilde{E}$.

\begin{figure}[t]
    \centering
    \includegraphics[width = 1\columnwidth]{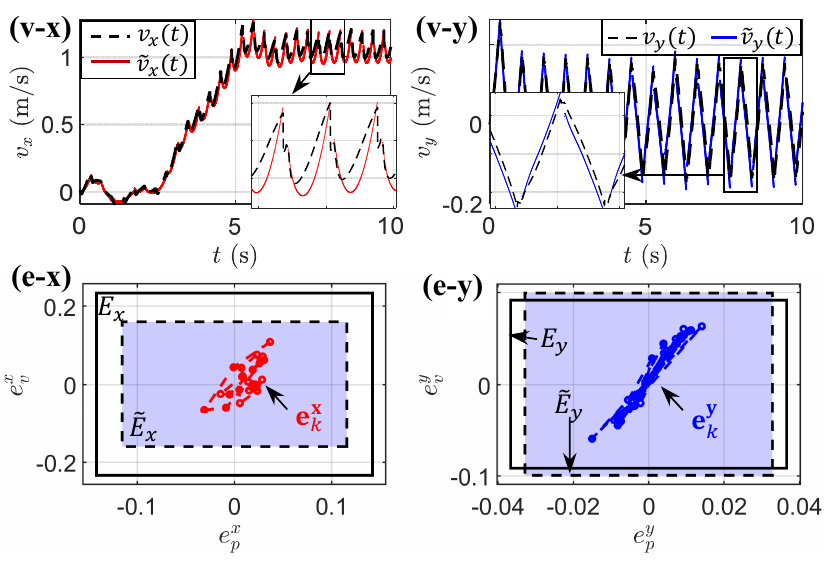} 
    \caption{Validation on the H-LIP based velocity approximation on a simulated walking with $v^t_{x,y} = [1, 0]$m/s, ${u^y_\text{L}}^* = -0.2$m: (v-x, v-y) the approximated horizontal velocities of the COM $\tilde{v}_{x}(t)$ (red line) and $\tilde{v}_{y}(t)$ (blue line) in the SSP compared with the actual velocities (dashed black lines); (e-x, e-y) the error state trajectories $\mathbf{e}^\mathbf{x,y}$ and the new error invariant sets $\tilde{E}_{x,y}$ (blue transparent polytopes) in each plane compared with the error invariant sets $E_{x,y}$ (white polytopes with black continuous bounding lines) from Fig. \ref{fig:simForwardRobotVsLIP}. }
    \label{fig:expVelApprox}
\end{figure}


To validate this, we use the H-LIP based velocity approximations to replace the actual horizontal velocities of the COM in the controller in simulation. The performance is comparable with that using true COM velocity in the previous section. Fig. \ref{fig:expVelApprox} shows the results on a simulated forward walking as a proof. As the horizontal COM dynamics of the robot is close to the H-LIP dynamics, the velocity approximation works well.  Although the new disturbance $\tilde{w}$ is $w$ plus another term, it does not necessarily mean that the size of $\tilde{W}$ and the resultant $\tilde{E}$ are larger. Here we get a smaller set in the sagittal plane (Fig. \ref{fig:expVelApprox} (e-x)), and the sets in the coronal plane are of similar sizes (Fig. \ref{fig:expVelApprox} (e-y)). \rev{Note that the H-LIP based velocity approximation is not designed as a general framework to replace any principled state estimators for legged robots but is shown to work efficiently and effectively along with our walking gait synthesis and hardware limitations.}

\subsubsection{Joint-level Controller} The optimization-based controller in section \ref{sec:Cassie} can be potentially implemented on the hardware by utilizing the secondary computer on the robot. Here we apply a \textit{PD + Gravitation Compensation} (PD+G) controller, which in practice provides an equivalent tracking performance and a much-lower computational effort. The PD+G controller is directly implemented on the main computer on the robot, which is written as:
$\tau_m = \tau_\text{PD} + \tau_\text{G},    
$
where $\tau_\text{PD}$ represents the PD component and $\tau_\text{G}$ represents the gravitation compensation part. 
 
For the PD component, we directly map the desired acceleration of the output $\ddot{\mathcal{Y}}^d$ to the joint torques. $\ddot{\mathcal{Y}}^d$ is identically chosen to that in \eqref{eq:outputLinearTSC}. $\mathcal{Y}$ and $\dot{\mathcal{Y}}$ are measured on the hardware. Note that the output selections (e.g. \eqref{eq:cassieOutputSSP}) are mainly functions of the motor joints. The actual acceleration of the output is assumed to be:
 $
 \ddot{\mathcal{Y}} = J_\mathcal{Y} \ddot{q}_m + \dot{J}_\mathcal{Y} \dot{q}_m,
$
 where $J_\mathcal{Y} = \frac{\partial{\mathcal{Y}}}{{\partial q_m}}$. The desired accelerations of the motor joints are applied as the motor torques:
$
    \tau_\text{PD} = \ddot{q}^d_m = J_\mathcal{Y}^{-1} (\ddot{\mathcal{Y} }^d - \dot{J}_\mathcal{Y}  \dot{q}). 
$

 For the gravity compensation, we need to find joint torques to cancel the gravitational terms in \eqref{eq:staticEOM} based on the current configuration and contact. The problem is inverse to the contact detection. Given $q$, we find $\tau_\text{G}$ to minimize:
$ \| B \tau_\text{G} + J^T_s \tau_s + J_h^T F_h - G(q) \|^2. $
 Note that the foot contact of the robot is underactuated and thus there does not exist any set of joint torques to completely cancel out the gravitational term unless the foot is fully-actuated. This yields a least square problem of
$
   \texttt{min}: \| \mathbf{A} \mathbf{X} - \mathbf{b} \|^2  
$
 where $\mathbf{A} = [B_m, J_h^T]$, $\mathbf{X} = [\tau_G^T, F_h^T]^T$, $\mathbf{b} = G(q) - J^T_s \tau_s$. Similarly, this problem can be solved via the pseudo-inverse of $\mathbf{A}$, i.e., $\mathbf{X} = \texttt{pinv}(\mathbf{A}) \mathbf{b}$, which yields the gravity compensation term $\tau_\text{G}$.
 
 \begin{figure}[t]
    \centering
    \includegraphics[width = 0.9\columnwidth]{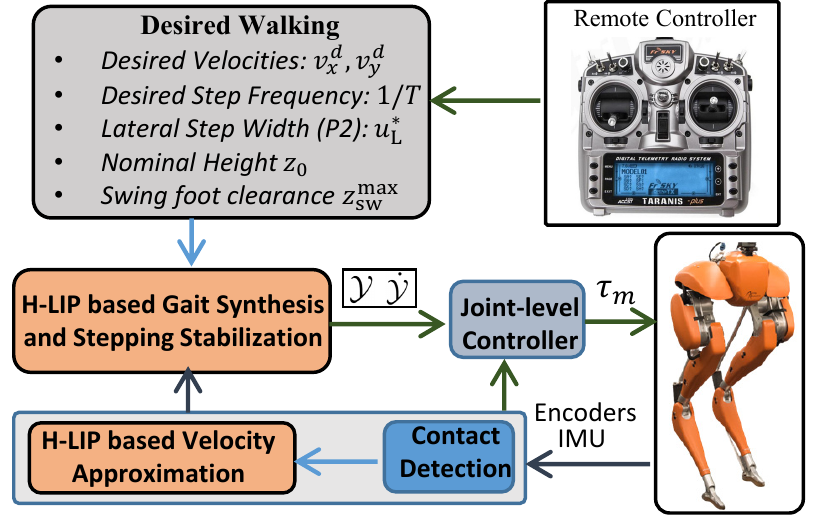}
    \caption{Illustration of the hardware realization of the controller on Cassie.}
    \label{fig:hardwareControl}
\end{figure}

\subsection{\rev{Hardware Implementation Scheme}}
The robot is controlled via a \rev{remote radio controller} that sends commands to the robot. The on-board computer is programmed to interpret the radio signals, read all the sensors on the robot, and send torque commands to the robot. The implementation is illustrated in Fig. \ref{fig:hardwareControl}. The \rev{remote radio} commands are processed to get the desired walking behaviors. \rev{The joysticks are used to provide desired walking velocities $v^d_{x, y}$ in the horizontal plane. We use low-pass filters to smooth the reading of the joysticks. Thus the desired velocities between consecutive steps do not vary significantly. The potentiometers on the remote controller are used to continuously select the gait parameters, e.g., desired step frequency, step width, nominal height, and swing foot clearance.} The H-LIP based gait synthesis and stepping calculate the output based on the gait parameters, contact, and COM states. The joint-level controller then calculates the motor torques and sends them to the motor modules to stabilize the outputs. \rev{We implement the controller on the on-board Intel NUC mini PC that runs Simulink real-time kernel. The control loop is set at 1kHz based on its computation capacity; further implementation optimization, e.g., on computation speedup, may be possible.}

In order to analyze the generated walking behaviors on the hardware, we use the EKF \cite{bloesch2013state} offline to get a continuous estimation of the horizontal velocity of the COM on the robot. The estimated velocities are used as references rather than the ground truth. \rev{We cross-validated the estimator with an external motion capturing system: velocity errors are within $0.1$m/s in the sagittal plane and $0.06$m/s in the lateral plane. In general, the estimation has non-neglectable errors that in nature come from the imperfections of the dynamics models and sensors. }

\subsection{Directional Walking}
\begin{figure}[t]
    \centering
    \includegraphics[width = .9\columnwidth]{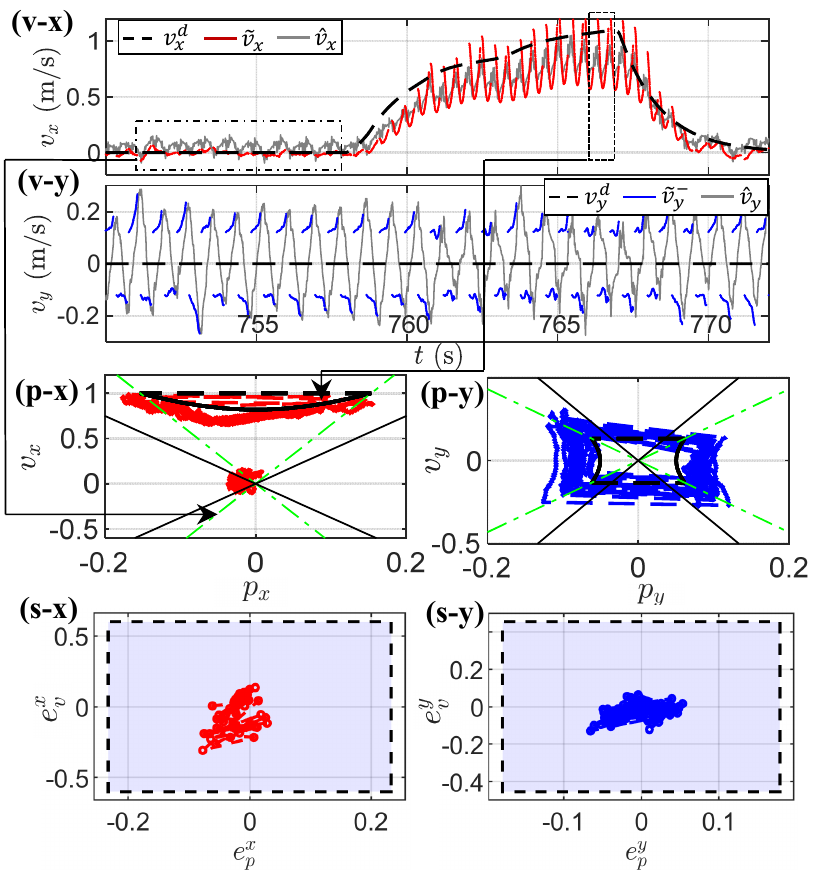} 
    \caption{Trajectories of a forward walking with varying target velocities. (v-x, v-y) plot the horizontal COM velocities including the desired velocities $v^d_{x,y}(t)$ (the black dashed lines), the velocity in the SSP from the H-LIP based approximation \rev{$\tilde{v}_{x,y}$} (the red lines), the predicted pre-impact velocity \rev{$\tilde{v}^-_{x,y}$}  (the blue lines), and the estimated velocities \rev{$\hat{v}_{x,y}$} (the gray lines). (p-x, p-y) plot the horizontal states in the sagittal (in different time segments) and coronal plane, respectively. The black orbits are the desired H-LIP orbits. (s-x, s-y) plot the error state trajectories (red and blue circles) inside the calculated error invariant sets $\tilde{E}_{x,y}$ (transparent blue polytopes) in each plane.}
    \label{fig:expForwardPlots}
\end{figure}

We demonstrate directional walking behaviors on the robot by using the joysticks on the \rev{remote controller} to steer the robot to its forward, backward, and lateral directions \cite{video:Directional}. Fig. \ref{fig:expForwardPlots} shows the horizontal COM states of a forward walking \cite{video:Directional}. The approximated velocities from the H-LIP are compared with the estimated velocities and the desired velocities. The desired walking velocities are tracked within reasonable errors. The error invariant sets are approximated in the same way (in Section \ref{sec:Sim}), and the error states are all inside the invariant sets. \rev{Note that the error invariant sets are larger than those in simulation in Fig. \ref{fig:expVelApprox}. This potentially is because the dynamics error $w$ is calculated using imperfect hardware data: $\mathbf{x}^R$ contains the measurement of $p^R$ and estimated velocity $\tilde{v}^R$ at the detected impact event, and $u^R$ is calculated from the kinematics and encoder readings. These errors directly yield larger sets $\tilde{W}_{x,y}$ and thus larger sets $\tilde{E}_{x,y}$.}

Additionally, the translational dynamics and transversal dynamics can be controlled separately. We thus implement a turning controller that only changes the hip yaw angles and keeps the stepping controller intact. With turning, the robot can be joystick-controlled easily in confined environments \cite{video:Directional}.

 \begin{figure}[t]
    \centering
    \includegraphics[width = 0.9\columnwidth]{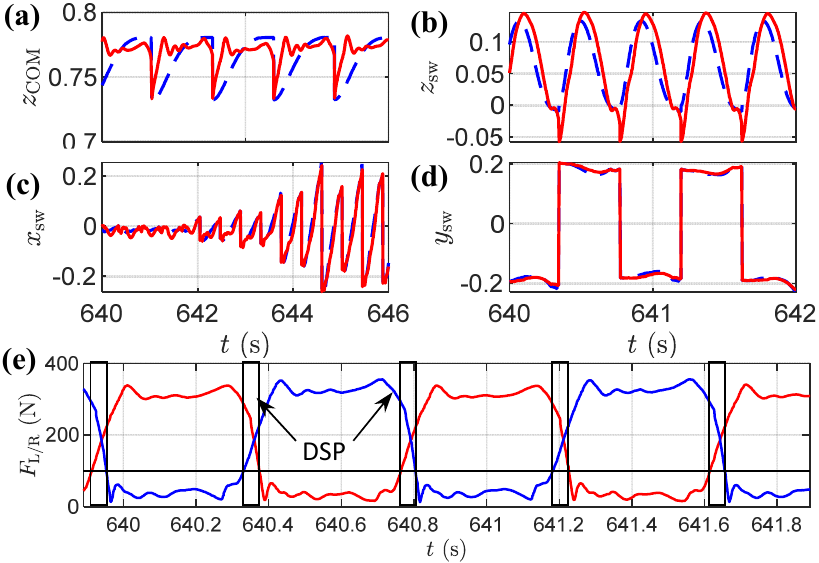}
    \caption{Illustrations of the output tracking and contact detection on the hardware: the desired output trajectories (the blue dashed lines) and the actual output trajectories (the red lines) of (a) the vertical COM position, and (b,c,d) the vertical, forward and lateral positions of the swing foot; (e) the contact detection via the GRF, where the boxed regions indicate the DSP.}
    \label{fig:expOutputs}
\end{figure}

\subsection{Versatile Walking}
\begin{figure}[t]
    \centering
    \includegraphics[width = .8\columnwidth]{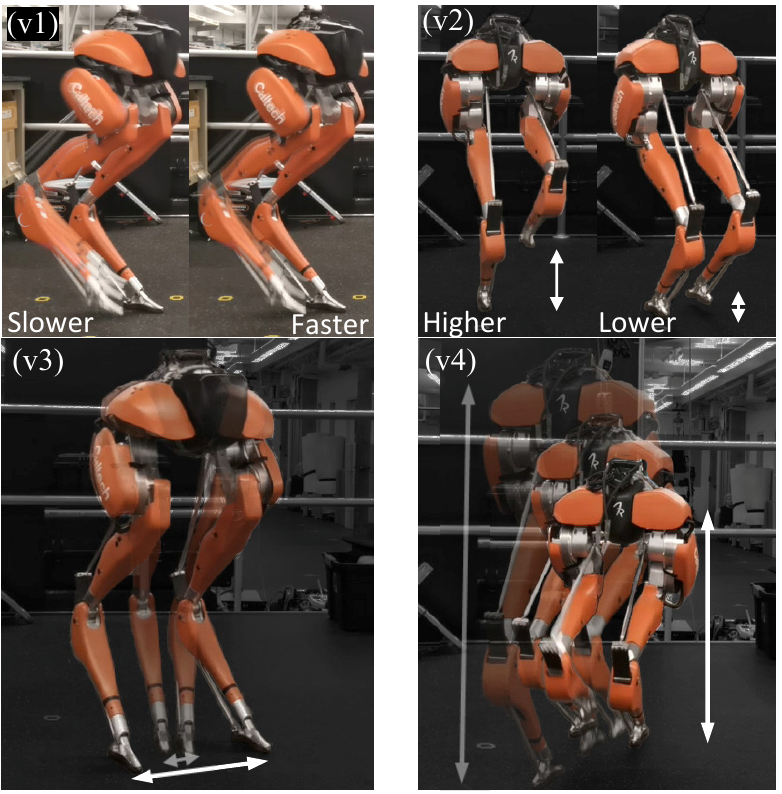}
    \caption{Illustration of the versatility of the realized walking by varying step frequency (v1), step clearance (v2), step width (v3), and COM height (v4).}
    \label{fig:versatile}
\end{figure}

\begin{figure}[t]
    \centering
    \includegraphics[width = .85\columnwidth]{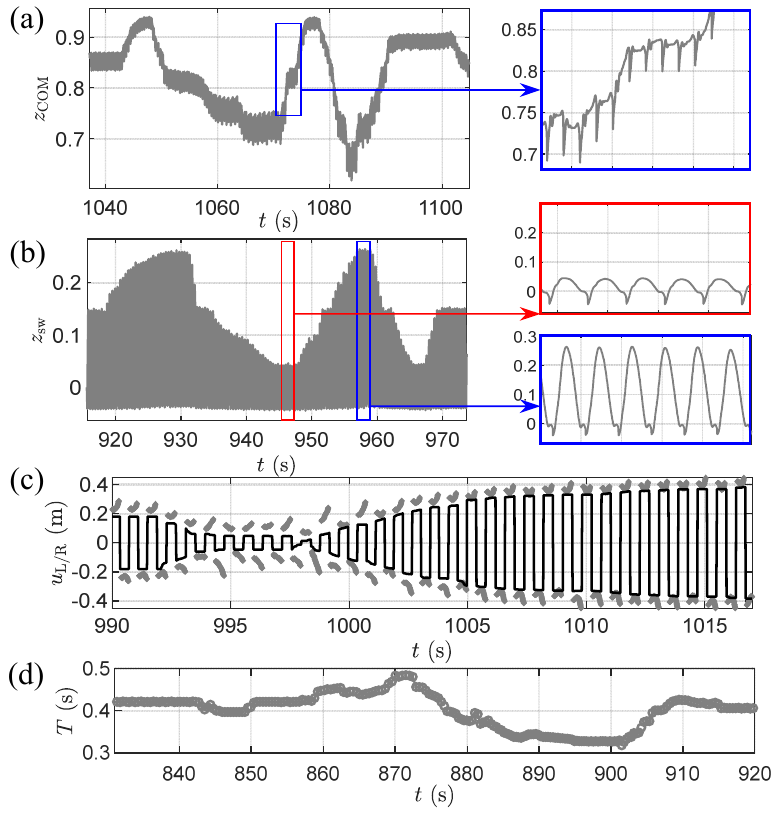}
    \caption{The trajectories of the (a) COM height and (b) vertical height of the swing foot in terms of the actual outputs, (c) the desired step width compared with the target step size $u^*_\text{L/R}$ (black lines), and (d) the duration of the walking.}
    \label{fig:versatilePlots}
\end{figure}

Now we utilize the potentiometers on the \rev{remote controller} to vary the gait parameters in real-time. 
The potentiometer readings can jitter, and we do not low-pass filter the values to show the robustness of our implementation. Fig. \ref{fig:versatile} demonstrates stepping-in-place with varying the four parameters, the ranges of which are listed in Table \ref{tab:versatile}. All the parameters can be varied continuously, and the H-LIP based approach still can stabilize the walking. Fig. \ref{fig:versatilePlots} demonstrates the continuous changes of the values in the experiment \cite{video:versatile}. Additionally, \cite{video:versatile} shows forward walking behaviors with different COM heights.

\begin{table}[!t]
    \caption{Versatile Walking Parameters} 
    \centering
\begin{tabular}{|c|c|c|}
\hline
 \textbf{Variables} & \textbf{Definition} & \textbf{Range} \\ \hline \hline
   Step Duration      &  $T$  & $0.3 - 0.5$s\\ \hline
  Desired Swing Foot Clearance      & $z^\text{max}_\text{sw}$ & $0.04 - 0.25$m  \\ \hline
  Desired Step Width &  $u^*_\text{L}$  & $0.08-0.45$m \\ \hline 
  Desired COM Height & $z_0$ &   $0.5 - 1$m \\ \hline
    \end{tabular}
    \label{tab:versatile}
\end{table}

\rev{These parameters do not strictly include all the admissible ones for the robot. Complete characterization of the admissible ranges of parameters has to explore the high-dimensional state space of the robot on its kinematic and dynamical feasibilities under a set of chosen control gains, which is inherently challenging; gait parameter ranges are also conditionally-coupled, e.g., a tall COM height provides a larger range of the swing foot clearance. Empirically speaking, for parameters outside of the ranges in the table, the walking can be kinematically/dynamically infeasible or destabilized under the current controller. For instance, the step width can not be too small or too larger, and the desired COM can not be too tall or too low simply due to the joint limits on the robot.} If the foot clearance is extremely low, the robot then has a trivial SSP and can not stabilize its walking via stepping; if it is too high, the vertical swing trajectory then requires large accelerations to lift-off and touch-down and thus exceeds the joint actuation limits. Similarly, the actuation limits prevent $T_\text{SSP}$ from being too small to track the swing trajectories. If the duration is too long, the robot can fall over \rev{($p^R$ exceeds its feasible range and $z_\text{COM}$ thus decreases)} before the swing leg strikes the ground to stabilize it.

The changes of the COM height and the step duration change the S2S dynamics of the H-LIP (e.g. \eqref{eq:HLIP_S2S}). The H-LIP stepping directly responds to the new S2S dynamics. Note that the vertical COM height is assumed constant on the H-LIP and that of the robot is controlled approximately constant. The height, however, can change between steps, as long as the vertical dynamics is not causing significant disturbance to the horizontal dynamics. The change of the swing foot clearance can change the impact velocity and potentially change $w$. Similarly, the step frequency variation changes the integration of the dynamics error in the continuous domains, which then change $w$. In the experiment, the qualitative and quantitative effects of these parameters on $w$ and then $E$ are not analyzed due to the existence of the horizontal velocity error (from the H-LIP based velocity approximation in the control or the state estimation in the analysis). Instead, the experiment shows that versatile walking behaviors are stably generated with the parameter variations on the fly.

\subsection{Disturbance Rejection}
Lastly, we demonstrate the robustness of the walking controller on the hardware \cite{video:robust}. Since the H-LIP stepping provides \textit{COM state-dependent step size planning} in \eqref{eq:robotDesiredStepSize}, the robot \rev{instantaneously and constantly} reacts to external disturbances. We consider two types of disturbances: external pushes and ground variations. The external pushes directly disturb the S2S dynamics of the robot; \rev{the ground variations change the domain durations, contact conditions, and vertical COM behaviors, which indirectly disturb the horizontal S2S dynamics.}

\begin{figure}[t]
    \centering
    \includegraphics[width = 1\columnwidth]{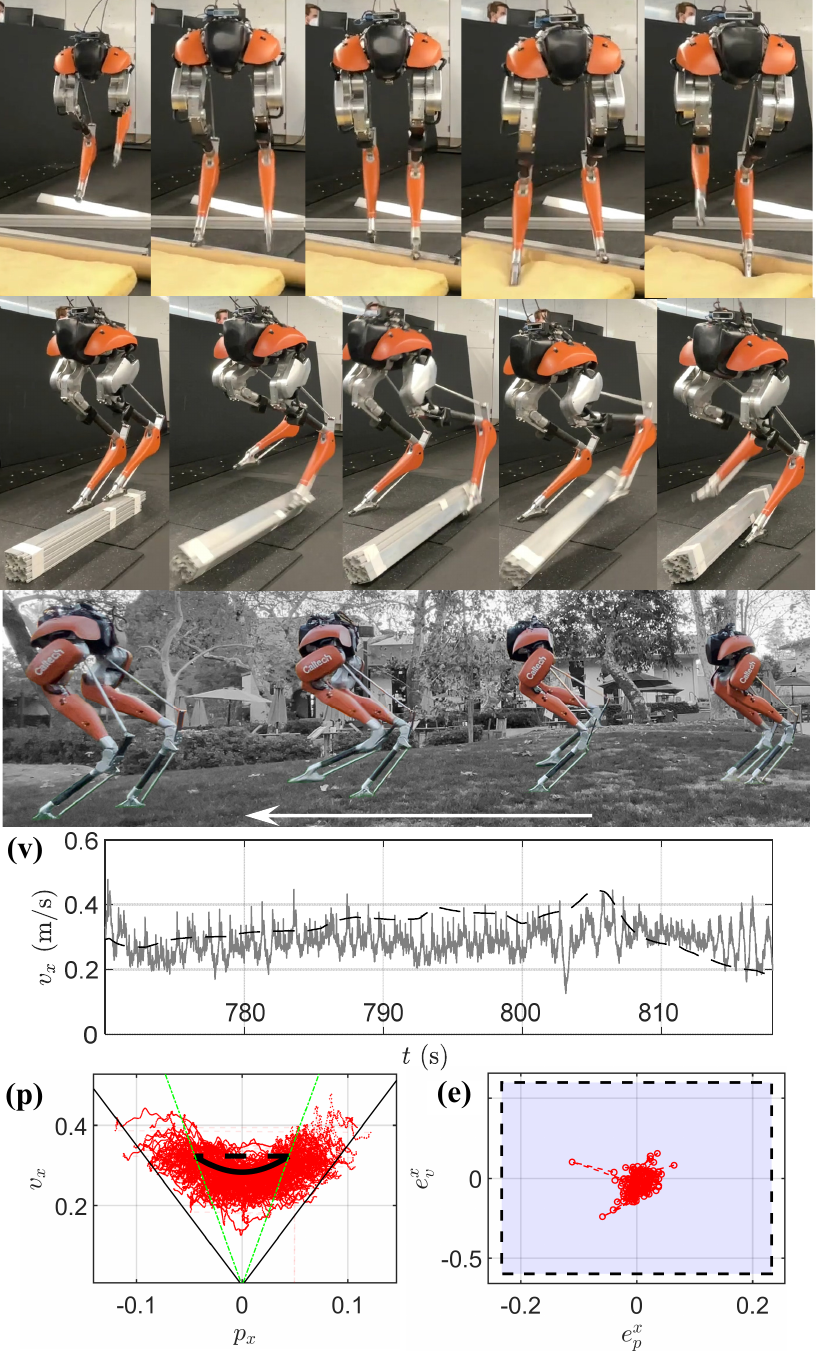}  
    \caption{\rev{Indoor and outdoor experiments on walking with terrain variations: on cluttered and uneven floor, an unstable obstacle, and grassy and uncertain terrain.} Analysis on walking on grassy terrain: (v) the estimated forward velocity (gray line) and the desired velocity (black dashed line), (p) the horizontal state trajectory (the red) in the sagittal plane compared with the desired orbit of the H-LIP (the black) for a walking segment with an approximately constant $v^d_x$ of $0.3$m/s, and (e) the error state trajectory compared with the undisturbed error invariant set $\tilde{E}_x$. }
    \label{fig:cassieGrass}
\end{figure}

\begin{figure}[t]
    \centering
    \includegraphics[width = .9\columnwidth]{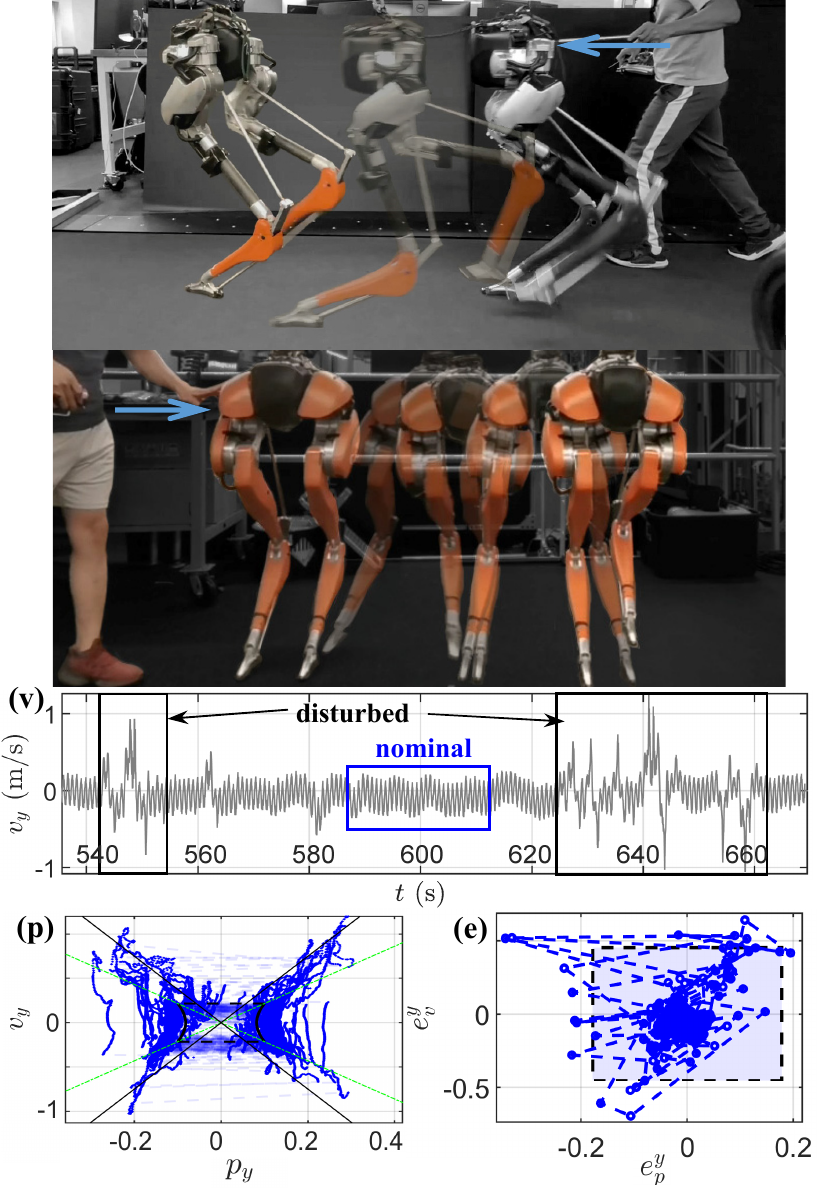}  
    \caption{\rev{Push recoveries on Cassie with analysis of the lateral pushes}: (v) the estimated horizontal velocity, (p) the horizontal state trajectory (the blue) compared with the desired orbit of the H-LIP (the black), and (e) the error state trajectory compared with the undisturbed error invariant set $\tilde{E}_y$.  }
    \label{fig:cassiePush}
\end{figure}

\rev{Both disturbances perturb the hybrid walking dynamics in very complex ways. It is inherently difficult to quantitatively characterize how the external disturbances affect the continuous dynamics. The S2S dynamics however can be concisely represented in the same form as: 
\begin{equation}
    \mathbf{x}^R_{k+1} = A \mathbf{x}^R_{{k}}  +  B u^R_{k} + w_k + w^\text{external}_k,
\end{equation}
where $w^\text{external}_k$ represents the influence of the external disturbances to the S2S dynamics of the horizontal COM states. Applying the H-LIP based stepping ($u^\rev{R} = u^{\text{H-LIP}} + K (\tilde{\mathbf{x}}^\rev{R}  - \mathbf{x}^{\text{H-LIP}})$) yields the error S2S dynamics: 
\begin{equation}\mathbf{e}_{k+1} = (A+BK) \mathbf{e}_{k}  + \tilde{w}_k + w^\text{external}_k,
\end{equation}
where $\tilde{w} = w + B K \delta \mathbf{x}_k$ is defined in \eqref{eq:errorDynUnderVelApprox} with $\delta \mathbf{x}_k$ being the "measurement error" on the horizontal COM state from the H-LIP based velocity approximation under the external disturbance. Since the external disturbance can not be characterized as a \textit{priori}, the goal of the stepping controller is simply to keep $\mathbf{e}$ small. When the external disturbances are bounded, $w^\text{external}$ and $\delta \mathbf{x}_k$ (and thus $\tilde{w}$) are bounded. Then the error state could potentially live inside the undisturbed $\tilde{E}$ (when $w^\text{external}$ are small) or converge to a different set (when $w^\text{external}$ are large). However, on the physical robot, under excessive disturbances, $\mathbf{e}$ grows quickly, and it is possible that the desired step sizes $u^d$ from the deadbeat controller can not be realized due to the kinematic feasibility ($u^R\in U$) or dynamic feasibility on swing foot tracking; the robot state $\mathbf{x}^R = \mathbf{x}^\text{H-LIP} + \mathbf{e}$ could exceed its feasible set $X$ due to joint limits, and the robot could no longer well-track the outputs, e.g., remain its vertical COM height, which leads to falls. 

Fig. \ref{fig:cassieGrass} demonstrates the walking with terrain variations that include both indoor and outdoor experiments where the robot walks blindly without using external sensors. The operator only uses the joystick to provide desired walking directions and velocities. Although the terrain could deform, roll, elevate, drop, or become unstable, the controller was able to balance the robot and track the desired walking by reactively placing the foot to reject the terrain disturbances. The continuous horizontal velocities have large variations (partially from the estimator) as shown in Fig. \ref{fig:cassieGrass}; the error states could lie inside $\tilde{E}$ under mild terrain variations. 

Fig. \ref{fig:cassiePush} shows the results on push recoveries in both sagittal and lateral plane. The operator sporadically applies horizontal push forces on the robot. The desired behavior is stepping-in-place. As expected, the error state $\mathbf{e}$ can temporarily go outside of the undisturbed invariant set $\tilde{E}$. The stepping controller then brings $\mathbf{e}$ back in $\tilde{E}$. In terms of the horizontal velocity, the robot is pushed to have large velocities and then the stepping controller drives the robot back to its nominal walking behavior. If the push is excessive, as is predicted, the robot could fall over due to its kinematic/dynamic incapability to realize the desired $u^d$ or remain upright with the resulting $p^R$. Since the robot is designed with larger kinematic ranges in its sagittal plane than the lateral plane, it can resist larger pushes in the sagittal plane \cite{video:robust}. Although the deadbeat stepping controller is not yet synthesized to maximize robustness, the experiments demonstrate the controller successfully rejecting reasonably-large external disturbances.

}

\section{\rev{Conclusion and Discussion}}
\label{sec:discuss}
To conclude, we present a Hybrid Linear Inverted Pendulum (H-LIP) based approach to control foot-underactuated bipedal walking. Periodic orbits of the H-LIP are comprehensively characterized in its state space and then orthogonally composed for 3D walking. The walking behaviors of the H-LIP are then approximately realized on the physical robot via the H-LIP based gait synthesis and stepping stabilization. The implementation is straightforward and computationally efficient. There are no non-convex optimizations to be solved offline or online. The realized walking behaviors are demonstrated to be both highly versatile and robust. 

We next discuss the implications, limitations, extensions, and future directions of the approach as the end. 
\subsection{Implications}
\subsubsection{Approximated Analytical Continuous "Gait Library"}
The orbit characterization of the H-LIP can be viewed as providing an approximated analytical "gait library" for the horizontal COM states of the bipedal robot. The "gait library" of the H-LIP is continuous, i.e., filling the state-space of the horizontal COM. Although the horizontal COM of the robot does not necessarily behave identically to the orbit, it converges closely to the orbit under the H-LIP stepping. More importantly, transitions between "gaits" or non-periodic walking behaviors can be easily realized via the H-LIP stepping. 

\subsubsection{Gait Synthesis and Characterization} 
The 3D composition of planar orbits offers a way of synthesizing and characterizing 3D bipedal walking gaits. The gait synthesis and characterization via composition of planar orbits can potentially be extended to other multi-legged systems, e.g., the bounding behavior on quadrupedal locomotion \cite{ding2020representation} can be viewed as producing a P2 orbit in its sagittal plane. The extension appears to be non-trivial but possible.  

\subsubsection{Model-free Planning}
The H-LIP based approach can be viewed as a "model-free" approach, where the robot model is not used in the planning. The walking of the H-LIP is shown to approximate the general hybrid nature of alternating support legs in bipedal walking. The planning on the hybrid dynamics of all the degrees of freedom (dofs) is encapsulated into the control on the horizontal dynamics of the COM; the individual dynamics of each dof is not specifically described. As a result, the approach can tolerate the imperfections of the robot modeling in the planning of walking. 

\subsubsection{Interpretation of Stability} The stability of underactuated bipedal walking is typically understood and analyzed on the periodic orbit of the robot \cite{grizzle2014models}. The S2S dynamics formulation provides a different perspective towards understanding the stability of walking. Assuming that the strongly-actuated dynamics (the outputs) can be stabilized, the underactuated/weakly-actuated dynamics (the horizontal COM states) are shown to be directly controlled by the step sizes in the S2S at the step level. Stabilization on the underactuated dynamics can thus be directly synthesized. The stability of the walking is no longer on the periodic orbits but on the viability of the discrete horizontal COM states in its feasible set.

\subsection{Limitations and Potential Solutions}

\subsubsection{Pelvis Orientation and Swing Foot Trajectory} The pelvis/upper-body orientation is fixed, and the swing foot trajectory is designed in the simplest way possible. Both are not optimized in terms of any criteria, e.g., energy consumption. It is possible to learn a low dimensional representation of the energy consumption in terms of parameterized trajectories of the swing foot or the pelvis. Optimal trajectories can then be constructed on the swing foot and the pelvis. 

\subsubsection{Performance Accuracy} The error state $\mathbf{e}$ directly describes the performance of the stepping controller which drives the robot to a desired walking of the H-LIP. The error is not controlled to zero but in the error invariant set $E$. There are two ways to further improve the performance in terms of reducing $\mathbf{e}$. The first is to develop a better approximation of the S2S dynamics so that the model difference $w$ is smaller, which will be further discussed later. The second is to employ a controller that can directly reduce the error $\mathbf{e}$; e.g., integral control is potentially able to mitigate the error. 

\subsubsection{Kinematic Feasibility} The robot joints are designed with limited ranges of motion, which limit the behaviors (i.e. the walking speeds and orbit compositions) on the robot. Additionally, the legs can internally collide with each other within their ranges of motion. This is more evident on Cassie due to its complex design. The stepping controller presented in this paper does not systematically take this into consideration. Instead, the kinematic feasibility is reflected on the choices of the desired walking of the H-LIP. In practice, this is sufficient to produce safe (despite conservative) walking on the robot; \rev{however, the feasibility is no longer guaranteed under external disturbances. Despite being very challenging, the kinematic feasibility should be identified systematically. Advanced optimization-based robust stepping controllers \cite{MAYNE2005219, xiong2021robust} can be explored to include the state and input bounds from the kinematic feasibility; the cost could be optimizing certain performance criteria such as walking speed or simply minimizing the error states, which can further increase the robustness to external disturbances to prevent falling.}

\subsubsection{Dynamic Feasibility} The realized walking behavior is assumed to be dynamically feasible; the desired trajectories of the outputs are assumed to be trackable given the hardware design. Optimization-based controller includes the torque bounds. However, theoretically, it does not guarantee the trajectories (especially the swing foot) to be well-tracked, e.g., when the walking duration is too small, the motors may not be able to move fast enough to drive the swing foot to the desired location. In practice, this can be identified empirically on the hardware despite the loss of theoretical soundness. 

\subsubsection{Vertical COM} The vertical COM is controlled approximately constant in each step. It permits gradual variations of the COM height between steps. It is not yet known if it could dramatically change the COM height within a step. One possible solution to enable this is to employ a model (e.g. a height-varying pendulum \cite{CaronVaryingHeight, koolen2016balance}) that captures both the vertical and horizontal COM behaviors. \rev{Varying vertical COM also alters the kinematic range of $p^R$ and $u^R$ and the horizontal S2S dynamics, which potentially could increase the walking robustness to external disturbances.} 

\subsection{Extensions and Future Directions}
\subsubsection{Global Position Control}
The H-LIP based stepping can also be used for controlling the global position \cite{gao2019global} of the underactuated bipedal robot \cite{xiong2021icra, Reza2021icra} by including the global position in the S2S dynamics. Then the H-LIP stepping can be used to approximately control the global position of the robot, where the feedback is on the error in terms of the global horizontal position, local horizontal position (w.r.t. the stance foot), and the horizontal velocity of the COM. 

\subsubsection{Walking over Rough Terrain} The H-LIP and the robot are assumed to walk on flat terrain here. The walking synthesis is shown to stabilize the robot walking on mildly uneven terrain. The H-LIP based approach can also be rigorously extended to walk on stairs, slopes and general rough terrains \cite{xiong2021ral}. A linear S2S dynamics approximation can be obtained if the vertical COM position is controlled with an approximately constant height from the ground. However, it is non-trivial to control the vertical COM of a compliant robot to follow certain desired trajectories, which will be one of the future work. 

\subsubsection{Improving S2S Approximation}
The S2S dynamics of the H-LIP is a linear model-free approximation and renders closed-form controllers for stabilization. The S2S approximation can potentially be improved, e.g., different dynamics quantities such as the angular momentum \cite{gong2020angular, dai2021bipedal} can also be explored for improvement. It is also possible to investigate data-driven approaches (e.g. \cite{bhounsule2020approximation, morimoto2007improving, xiong2021robust}), which could potentially offer better approximations and thus improve the performances on the stepping stabilization.

\subsubsection{On Fully-actuated Humanoid Walking} 
The H-LIP based approach can also be potentially applied towards walking on fully-actuated humanoids. The foot is then actuated but with limited controls, which comes from the ankle actuation and zero moment point (ZMP) constraint on the support polygon. The foot actuation helps to control the robot. Therefore, in the future, we will explore the integration with H-LIP based approach and the foot actuation on humanoids for generating highly dynamic and versatile behaviors.

\section{Appendix: proofs on the H-LIP}

\noindent{\underline{\textit{Proof of Thm. 1:}}}
Combining \eqref{eq:SSPsol} and \eqref{eq:ImpactS2S} yields \eqref{eq:sigma1}, and $p^+ = -p^-$, plugging which into \eqref{eq:ImpactS2S} yields \eqref{eq:u*P1}.

\noindent{\underline{\textit{Proof of Thm. 2:}}}
We can first show that, any state on the line $v = - \sigma_2 p + d_2$ in the beginning of the SSP will flow to the line $v = \sigma_2 p + d_2$ at the end of the SSP, with $d_2$ being a constant. This is easily proven by using \eqref{eq:SSPsol}. Then an arbitrary state is chosen, and it is easy to show the step sizes must be $u_\text{L/R} =2 p^{-}_\text{L/R} +  T_{\textrm{DSP}} v^{-}_\text{L/R}$ to get a two-step orbit. 

To derive $d_2$ from the desired velocity, we first select an arbitrary state $[p_0, - \sigma_2 p_0 + d_2]$ as the initial state of the P2 orbit. The rest of the boundary states can be calculated as functions of $p_0$ and $d_2$. Then the sum of the step sizes is
$u_\text{L}^* + u_\text{R}^* = d_2 (T_\text{DSP} + T_\text{DSP} \text{cosh}(T_\text{SSP} \lambda) + \frac{2}{\lambda} \text{sinh}( T_\text{SSP} \lambda))$ which is equal to $2 v^d T$. Solving this for $d_2$ yields \eqref{eq:d_2}. 

\noindent{\underline{\textit{Proof of Prop. 3:}}} This proof follows the previous paragraph by starting an arbitrary state $[p_0, - \sigma_2 p_0 + d_2]$ as the initial SSP state of the P2 orbit. Letting $u_\text{L} = u_\text{R}$ and solving for $p_0$ yields $p_0 = -\frac{d_2 \text{sinh}(T_\text{SSP} \lambda)}{2\lambda}$. The rest follows immediately. 

\noindent{\underline{\textit{Proof of Prop. 4:}}}
This proof is similar to the proof of Them. 2. First, it is easy to show that any initial state on the line $v = - \sigma_1 (p + d_1)$ will flow to the line $v = \sigma_2 (p + d_1)$ after $T_\text{SSP}$. Similarly, any initial state on the line $v = - \sigma_1 (p - d_1)$ will flow to the line $v = \sigma_2 (p - d_1)$ after $T_\text{SSP}$. Then an arbitrary state is chosen on the line $v = \sigma_2 (p + d_1)$ (or equivalently $v = \sigma_2 (p - d_1)$), and it is easy to show the step sizes must be $u_\text{L/R} =2 p^{-}_\text{L/R} +  T_{\textrm{DSP}} v^{-}_\text{L/R}$ to get a two-step orbit.

\section{Appendix: Bézier Polynomials}
The desired output trajectories are designed via Bézier polynomials:
$
    b(t) = \mathcal{\beta} (\bar{t}(t):= \frac{t}{T}): =  \sum_{k=0}^{M} \beta_k \frac{M!}{k!(M-k)!}\bar{t}^k(1-\bar{t})^{M-k}, \nonumber
$
where $\bar{t} \in [0,1]$ and $\beta_k$ are the Bézier coefficients. The following table lists the coefficients
, where $\mathbf{1}_N$ indicates a row vector of size $N$ with all elements being 1.  
\begin{table}[h]
   \centering   
    \label{tab:bezier}
    \begin{tabular}{|c|c|c|}
    \hline
    On & Notation & Bézier Coefficients $\beta$\\
    \hline \hline
      Horizontal swing foot   &  $b_h$ &  $[0,0,\mathbf{1}_3]$\\ \hline
      Vertical swing foot     & $b_v$ & $[0, z^\text{max}_\text{sw} \mathbf{1}_4, 0, z^\text{neg}_\text{sw}]$\\ \hline
      Vertical COM height     & $b_h$ &  $[0,0,\mathbf{1}_3]$\\ \hline
    \end{tabular}
\end{table}

\bibliographystyle{IEEEtran}
\bibliography{main}

\begin{IEEEbiography}[{\includegraphics[width=1in,height=1in,clip,keepaspectratio]{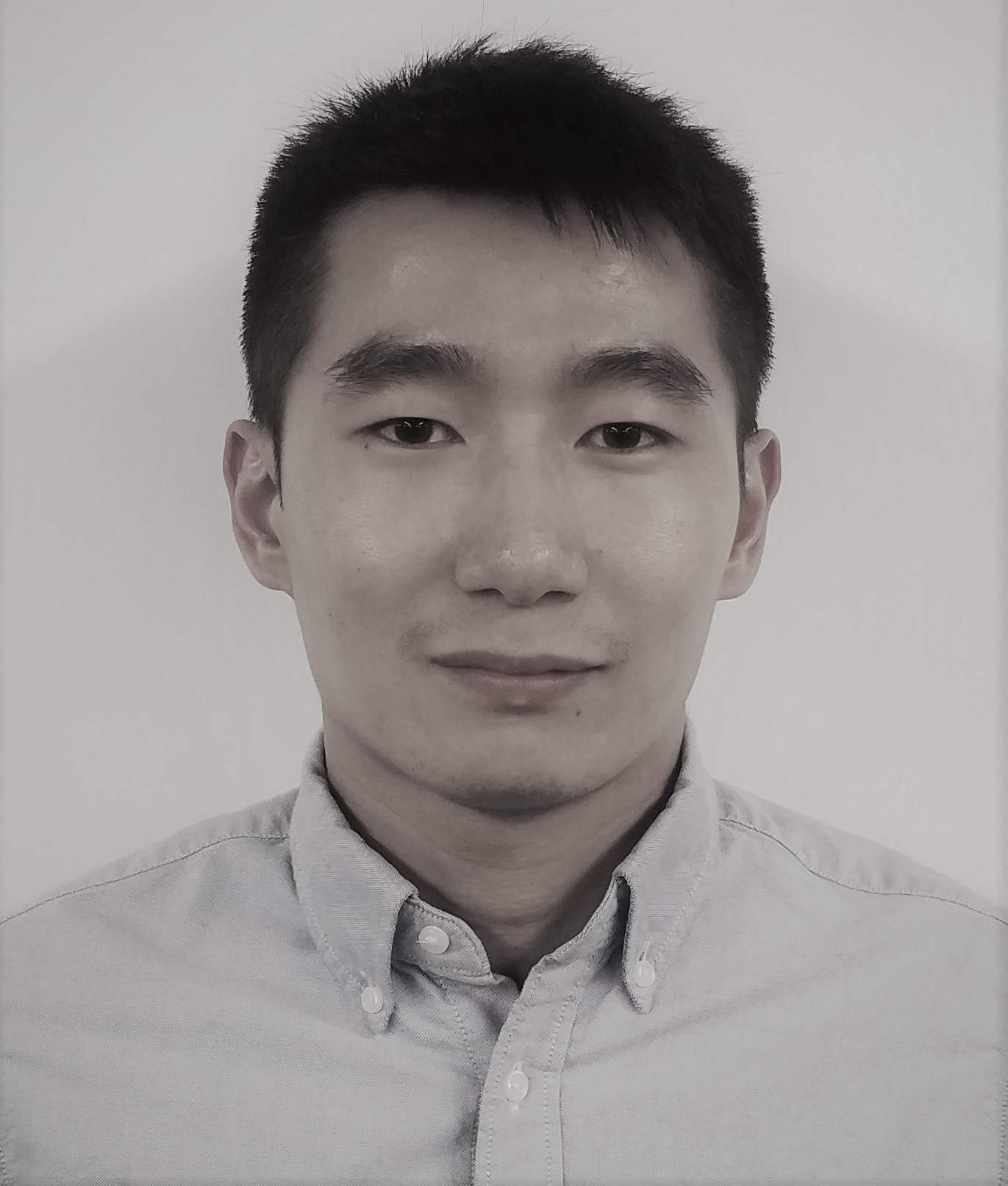}}]{Xiaobin Xiong} is currently a post-doc scholar at the California Institute of Technology (Caltech). He received his B.S. from Tongji University, Shanghai, China in 2013, and his M.S. from Northwestern University, Evanston, Illinois in 2015, and his PhD from Caltech in 2021, all in mechanical engineering. At Northwestern, he worked with Dr. Kevin M. Lynch and Dr. Paul Umbanhowar on nonprehensile manipulation. 
His research interest lies in controls and planning for robotic dynamical systems. He was one of the Amazon Fellows in AI at Caltech, and the recipient of the IROS-Robocup Best Paper Award in 2019.
\end{IEEEbiography}

\begin{IEEEbiography}[{\includegraphics[width=1in,height=1in,clip,keepaspectratio]{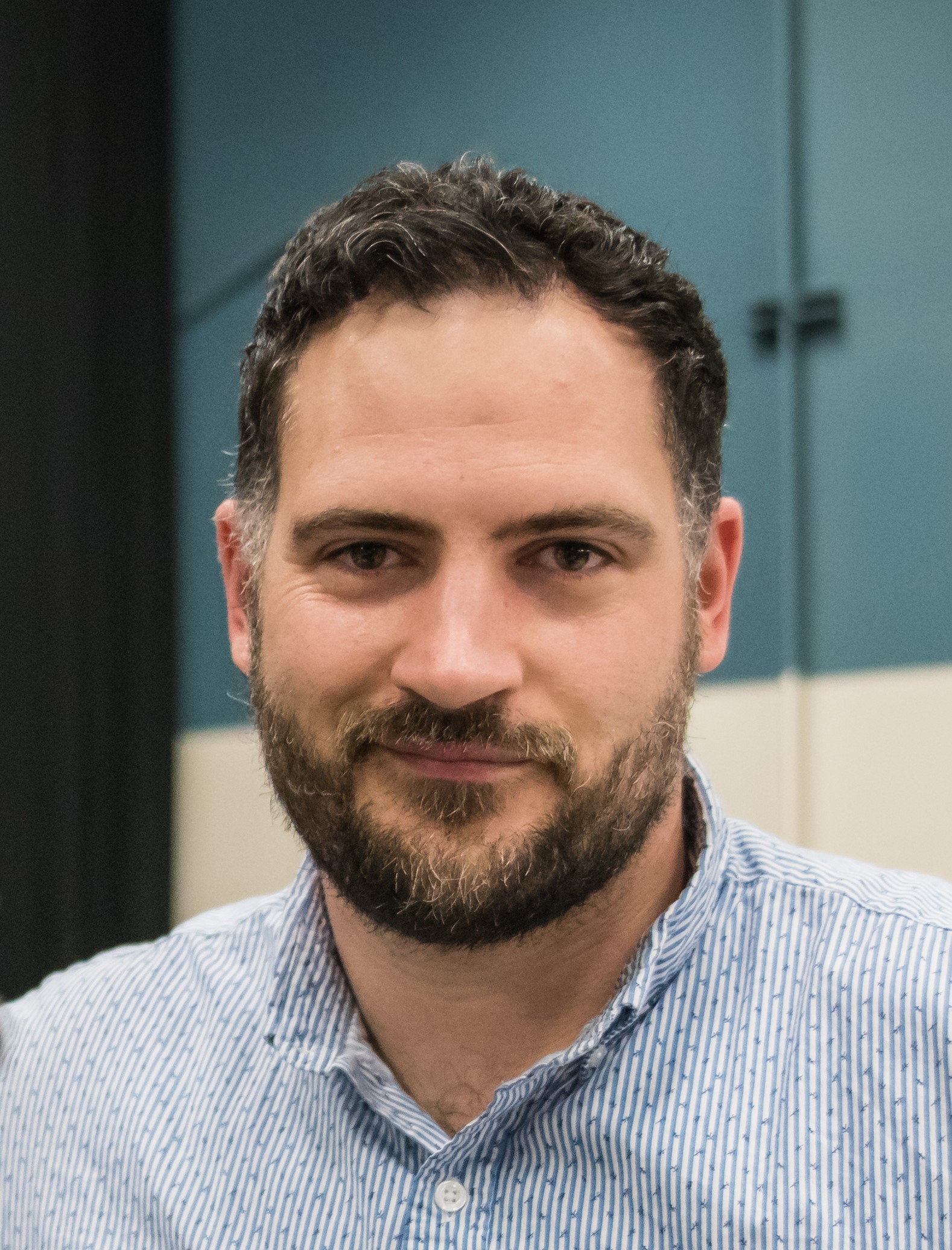}}]{Aaron Ames} (S'04–M'16-F'21) received a B.S. in mechanical engineering and a B.A. in mathematics from the University of St. Thomas in 2001, and he received a M.A. in mathematics and a Ph.D. in electrical engineering and computer sciences from UC Berkeley in 2006. He is the Bren Professor of Mechanical and Civil Engineering and Control and Dynamical Systems at Caltech. Prior to joining Caltech in 2017, he was an Associate Professor at Georgia Tech in the Woodruff School of Mechanical Engineering and the School of Electrical \& Computer Engineering. He served as a Postdoctoral Scholar in Control and Dynamical Systems at Caltech from 2006 to 2008, and began is faculty career at Texas A\&M University in 2008. His research interests span the areas of robotics, nonlinear control and hybrid systems, with a special focus on applications to bipedal robotic walking both formally and through experimental validation. His lab designs, builds and tests novel bipedal robots, humanoids and prostheses with the goal of achieving human-like bipedal robotic locomotion and translating these to robotic assistive devices. \\
Dr. Ames received the 2005 Leon O. Chua Award for achievement in nonlinear science, the 2006 Bernard Friedman Memorial Prize in Applied Mathematics, the NSF CAREER award in 2010, the 2015 Donald P. Eckman Award, and the 2019 IEEE CSS Antonio Ruberti Young Researcher Prize.
\end{IEEEbiography}

\end{document}